\PassOptionsToPackage{dvipsnames}{xcolor}

\documentclass[accepted]{uai2025} 
                        
\usepackage[american]{babel}

\usepackage{natbib} 
\usepackage{mathtools} 
\usepackage{booktabs} 
\usepackage{tikz} 

\usepackage{times}
\usepackage{latexsym}
\usepackage[T1]{fontenc}
\usepackage[utf8]{inputenc}
\usepackage{microtype}
\usepackage{inconsolata}
\usepackage{graphicx}
\usepackage{subcaption}
\usepackage{booktabs}
\usepackage{multirow}

\definecolor{mplblue}{RGB}{31,119,180}
\definecolor{mplorange}{RGB}{255,127,14}

\usepackage{algorithm}
\usepackage{algorithmic}
\newcommand{\algorithmicbreak}{\textbf{break}}
\newcommand{\BREAK}{\STATE \algorithmicbreak}
\usepackage{enumitem}
\usepackage{amsthm}
\usepackage{amsmath}
\usepackage{amsfonts}
\usepackage{amssymb}

\newtheorem{definition}{Definition}

\DeclareMathOperator*{\argmax}{argmax}
\DeclareMathOperator*{\argmin}{argmin}

\title{Multi-group Uncertainty Quantification for Long-form Text Generation}

\author[1]{Terrance Liu}
\author[1]{Zhiwei Steven Wu}
\affil[1]{%
    Carnegie Mellon University
}
  
\begin{document}
\maketitle

\begin{abstract}

%
While past works have shown how uncertainty quantification can be applied to large language model (LLM) outputs, the question of whether resulting uncertainty guarantees still hold within sub-groupings of data remains open.
In our work, given some long-form text generated by an LLM, we study uncertainty at both the level of individual claims contained within the output (via \textit{calibration}) and across the entire output itself (via \textit{conformal prediction}).
Using biography generation as a testbed for this study, we derive a set of (demographic) attributes (e.g., whether some text describes a man or woman) for each generation to form such ``subgroups'' of data. 
We find that although canonical methods for both types of uncertainty quantification perform well when measuring across the entire dataset, such guarantees break down when examining particular subgroups.
Having established this issue, we invoke group-conditional methods for uncertainty quantification---\textit{multicalibration} and \textit{multivalid} conformal prediction---and find that across a variety of approaches, additional subgroup information consistently improves calibration and conformal prediction within subgroups (while crucially retaining guarantees across the entire dataset).
As the problems of calibration, conformal prediction, and their multi-group counterparts have not been extensively explored in the context of long-form text generation, we consider these results to form a benchmark for this setting.

\end{abstract}

\section{Introduction}

In recent years, researchers have developed stronger large language models that perform well on a variety of tasks across different domains \citep{touvron2023llama, bubeck2023sparks, anil2023palm}. 
However, as use of LLMs continues to grow, so do concerns over their tendency to hallucinate facts \citep{huang2023survey}. As a result, there is a growing need for methods that can reduce hallucinations \citep{manakul2023selfcheckgpt, zhang2023alleviating}, perform abstention \citep{yang2023alignment}, or provide correctness guarantees \citep{kumar2023conformal, mohri2024language, quach2023conformal}. Our work focuses on the latter---broadly speaking, uncertainty quantification of long-form large language model generations.

Concretely, given a set of claims produced by an LLM in response to some prompt, our goal is to provide a confidence score or uncertainty guarantee about the factual correctness of the output. We explore this problem in two settings: given a set of claims contained within some long-form prompt response, we \textbf{(1)} ensure factuality at the individual claim level and \textbf{(2)} provide uncertainty guarantees across the whole set of claims. We approach problem \textbf{(1)} via \textit{calibration}, in which one wishes to output a calibrated score for each claim, while for problem \textbf{(2)}, we apply \textit{conformal prediction} \citep{shafer2008tutorial}, selecting a subset of claims that---with high probability---are \textit{all} correct.

In contrast to existing works on uncertainty guarantees of long-form generations \citep{quach2023conformal, mohri2024language}, we make the observation that while these guarantees may be valid under the full data distribution, they may not still be valid within individual subgroups of the distribution. For example, generations describing local politicians may be more prone to error than generations concerning national leaders. We choose biography generation as a testbed for multi-group uncertainty quantification, arguing that this problem is well-motivated, given that bias within biography generation has long been studied \citep{de2019bias}. Having derived a set of subgroups using demographic information  (e.g., whether an LLM output describes a man or woman), we find that when evaluated with respect to such groupings, canonical methods for calibration and conformal prediction indeed exhibit significant biases.\footnote{Uncertainty can be epistemic and aleatoric, and sources of group bias can be categorized into either (or both) types. Our work, which focuses on atomic factuality in long-form generation, falls under epistemic uncertainty.}

Having established such issues for standard uncertainty quantification approaches, we shift our attention to understanding to what extent such biases can be corrected. To address this unmet need, we introduce methods quantifying uncertainty in long-form text generation that are valid not only across a full distribution of prompts (i.e., marginally) but also across identifiable subgroups of prompts (i.e., conditionally). Invoking \textbf{(1)} \textit{multi}calibration \citep{hebert2018multicalibration} and \textbf{(2)}
\textit{multivalid} conformal prediction \citep{jung2022batch}, we categorize methods into two styles: iterative ``patching'' and linear regressor-based algorithms. 

Our results demonstrate that for both problems \textbf{(1)} and \textbf{(2)}, multicalibration and multivalid conformal prediction techniques improve measures of uncertainty relative to standard (marginal) calibration and conformal prediction methods. This advantage holds \textbf{\textit{regardless}} of whether evaluation is conducted within groups or across the entire dataset. As the problems of calibration, conformal prediction, and their multi-group counterparts have not been extensively explored in the context of long-form text generation, we consider these results to form a benchmark for this setting.


\subsection{Related work}

\paragraph{Factuality in long-form LLM outputs.} Evaluating factuality for long-form generation \citep{min2023factscore, song2024veriscore, weilong, bayat2024factbench} is challenging: not only do generated outputs consist of many parts that must be scored individually, but also scoring each part requires prohibitively costly manual annotation. To make evaluation more tractable, \citet{min2023factscore} introduce \textsc{FActSCORE}, which converts any generation into a set of atomic facts (claims) that are then labeled as true or false. Using this evaluation metric, \citet{min2023factscore} test LLMs' abilities to generate biographies and find that their generations are pervaded with errors. 



\paragraph{Attaching confidence scores to LLM outputs.} While a natural method for producing an uncertainty estimate is to use a model's output probabilities directly as a confidence score \citep{achiam2023gpt}, it has been shown that model probabilities are not well calibrated \citep{guo2017calibration}. As a result, many works have recently proposed alternative methods for generating uncertainty scores that can then be used to refine or correct LLM outputs \citep{ wang2022self, xiong2023can, geng2024survey, fadeeva2023lm, vashurin2024benchmarking}. 
We highlight that such work is complementary to our line of work---rather than proposing an entirely new uncertainty score function, we focus on how one can better leverage existing scores to produce uncertainty guarantees.


\paragraph{Uncertainty quantification for LLMs.} 
We note that much of the prior work on multicalibration and multivalid conformal prediction are rooted in theory. Like \citet{detommaso2024multicalibration}, our work tries to bridge the gap between theoretical insights and practical problems today (i.e., LLM generations). However, while \citet{detommaso2024multicalibration} calibrate for correctness in question-answering, we are the first to apply multicalibration to claims decomposed from long-form text generation. Moreover, unlike \citet{detommaso2024multicalibration}, we consider uncertainty quantification in the form of conformal prediction. 

In addition, our work closely relates to \citet{mohri2024language}, which aims to provide high probability guarantees of factuality in long-form generation. In particular, \citet{mohri2024language} frame this problem as a nested conformal prediction problem, producing subsets of claims that achieve some marginally valid coverage (i.e., produce some generation that on average, contains a correct output with any user-specified probability). Our work, however, extends this problem to multivalid conformal prediction: we produce generations that are not only correct on average but are also conditionally correct across subgroups.

Finally, concurrent work by \citet{cherian2024large} also builds off this framework, but unlike our work, they introduce a new objective in which the goal is to instead guarantee that at least some given proportion of claims are retained. By applying a method proposed in \citet{gibbs2023conformal}, \citet{cherian2024large}'s algorithm can (optionally) condition on group membership. However, their experiments include only 5 (non-overlapping) groups that are derived from the same feature, while our work focuses on the more challenging setting in which examples can simultaneously belong to many groups.
\section{Preliminaries}


\subsection{Calibration}\label{sec:calibration}

We begin by defining calibration in context of factuality in open-ended text generation. Suppose we are given some $(X,Y) \sim \mathcal{D}$ where $X \in \mathcal{X}$ denotes some claim outputted by an LLM, while $Y$ is an indicator in which $Y=1$ when the claim is correct (and $Y=0$ otherwise). Suppose there exists some uncertainty score function $f: \mathcal{X} \to [0, 1]$ that measures confidence for the correctness of some input $X$ (with higher values denoting higher levels of confidence). Then a goal one may have when designing such a score function $f$ is to have that
\begin{equation}\label{eq:perfect}
    P_{\mathcal{D}} (Y=1 \mid f(X) = p) = p, \forall x \in \mathcal{X}
\end{equation}
In other words, the probability that some LLM output is correct is given exactly by $f$.

Calibration, then, defines a simpler, more tractable condition, in which instead of ensuring guarantees across all possible values of $f$, it ensures a guarantee over coarser, level sets $S_p(f)$: 
\begin{definition}\label{def:calibration}
(Calibration) A function $f$ is calibrated w.r.t $D$ if
\begin{equation*}
    \Delta_p(f) = 0, \forall p \in [0, 1]
\end{equation*}
where $\Delta_p(f)$ is the bias of $f$ for the p-th level set  $S_p(f) = \{f(x) = p\}$:
\begin{equation*}
    \Delta_p(f) = \mathbb{E}_{\mathcal{D}}[Y - f(X) \mid S_p(f)]
\end{equation*}
\end{definition}

Defining level sets is akin to dividing the output space of $f$ (i.e., $[0,1]$) into buckets. For example, one could round $f(X)$ to the nearest value in some predefined set of probabilities (e.g. $\{0, 0.5, 1.0\}$). One can view this definition of calibration as a desirable guarantee since it serves as a minimal condition for Equation \ref{eq:perfect}---any $f$ that satisfies \eqref{eq:perfect} must (at the very least) also be calibrated. We note that to evaluate calibration, we can consider the average squared calibration error (ASCE) of $f$.
\begin{equation}\label{def:asce}
    \textrm{ASCE}(f) = \mathbb{E}_P [\Delta^2_{P}(f)]
\end{equation}
The ASCE averages the squared bias across all level sets and is zero when $f$ is calibrated.

\paragraph{Multicalibration.} 

While calibration provides an already important and useful guarantee, it can often be insufficient in many real-world scenarios. For example, in the context of generating information about people, one maybe desire that $f$ is calibrated not only across all people, but also within subpopulations defined by demographic attributes like \textit{sex or gender}. Otherwise, it is possible that certain subgroups can still suffer from very high miscalibration, even when the score function is perfectly calibrated across $\mathcal{D}$. Ideally, one would hope to have guarantees while conditioning on as many subgroups in $\mathcal{X}$ as possible, both from the perspective of machine learning fairness as well as enhancing the likelihood of correctness in general. 

Multicalibration \citep{hebert2018multicalibration} was developed to provide accurate guarantees across overlapping subgroups (i.e., a sample can belong to many groups). Let $g: \mathcal{X} \to \{0, 1\}$ be a group function that evaluates to $1$ if $X$ belongs to some group. We study, then, the setting in which there exists of set of groups $\mathcal{G}$ that corresponds to our data domain $\mathcal{D}$. While the set of groups can be disjoint, the problem of multicalibration then becomes trivial in this case because one can simply split a dataset into disjoint sets that can then each be calibrated individually. Consequently, prior work typically considers the more interesting case where many intersecting groups comprise $\mathcal{G}$.

Given a group function $g$, we define group average squared calibration error (gASCE) as:
\begin{equation}\label{def:gasce}
    \textrm{gASCE}(f, g) = \mathbb{E}_P [\Delta^2_{p,g}(f) \mid g(X) = 1]
\end{equation}
where 
\begin{equation*}
    \Delta_{p,g}(f) = \mathbb{E}_{\mathcal{D}}[Y - f(X) \mid S_{p, g}(f)]
\end{equation*}
for $S_{p, g} = \{f(X) = p, g(x) = 1\}$.
In other words, gASCE conditions on both level sets and group membership. Finally, we have:
\begin{definition}\label{def:multicalibration}
(Multicalibration) A function $f$ is $\alpha$-multicalibrated w.r.t $D$ and a set of groups $\mathcal{G}$ if and only if
\begin{equation*}
    gASCE(f, g) < \frac{\alpha}{P_{\mathcal{D}}(g(X) = 1)},\forall g \in \mathcal{G}
\end{equation*}
\end{definition}

\subsection{Conformal Prediction}\label{sec:conformal}

In conformal prediction, the general goal is to produce some confidence set $\mathcal{T}(X)$ for some example $X$ such that this set marginally \textit{covers} the true label $Y$ with some target probability $1-\alpha$.
\begin{equation}\label{eq:conformal}
    P_{\mathcal{D}} (Y \in \mathcal{T}(X)) = 1 - \alpha
\end{equation}

The second part of our work follows the problem statement outlined in \citet{mohri2024language}. Unlike in calibration, where each claim contained in some long-form generation is treated individually, \citet{mohri2024language} instead define their problem in terms of pairs $(X, Y)$, where $X$ is some input prompt and $L(X) = Y \in \mathcal{Y}$ is the long-form generation outputted by a LLM $L$. Because $Y$ may or may not be supported by some reference ground truth $Y^*$,\footnote{in the case of FActScore \citep{min2023factscore}, "is $Y$ supported by Wikipedia?"}
\citet{mohri2024language} define factuality in terms of entailment operations $Y^* \implies Y$. Furthermore, they rewrite this relation as $Y^* \in E(Y) = \{ Y' \in \mathcal{Y}: Y' \implies Y \}$. This equivalent set notation, in other words, means that some reference ground truth $Y^*$ (e.g., a Wikipedia article in \citet{min2023factscore}) is contained in the set of possible texts $Y'$ that support all claims made in the LLM output $Y$. 

Given this notation, the goal is to find some uncertainty set $\mathcal{T}(L(X))$ s.t. $P_{\mathcal{D}} (Y \in \mathcal{T}(L(X))) = 1 - \alpha$. In the context of long-form text generation, this goal translates to taking as input the original LLM output $L(X)$ and producing a subset of claims $\mathcal{T}(L(X))$ such that with high probability, $1 - \alpha$, all remaining claims are factually correct. 

We note that to empirically measure such guarantees, one can use the \textit{coverage error} of $\mathcal{T}$ w.r.t the target error rate $\alpha$.
\begin{equation}
    | P_{\mathcal{D}} (Y \notin \mathcal{T}(X)) - \alpha |
\end{equation}
%

\paragraph{Multivalid Conformal Prediction.}

Similar to calibration, one may also desire group conditional coverage guarantees for intersecting groups. Known as \textit{multivalid conformal prediction} \citep{jung2022batch}, these guarantees are stronger than marginal conformal guarantees, holding also when conditioned on group membership. Using group functions $g$, as defined in Section \ref{sec:calibration}, full multivalid coverage can be written as the following: Given some set of groups $\mathcal{G}$, we have that 
\begin{equation}\label{eq:multi_conformal}
    P_{\mathcal{D}} (Y \in \mathcal{T}(X) \mid g(X) = 1) = 1 - \alpha
\end{equation}
for all group functions $g \in \mathcal{G}$. Thus, target coverage guarantees $1-\alpha$ must hold both marginally and within all subgroups.

\section{Methods} \label{sec:methods}

Next, we introduce the methods (and their group-conditional variants) for applying calibration and conformal prediction to language model factuality. We organize these methods into two categories: \textbf{(1)} iterative ``patching''-based algorithms and \textbf{(2)} linear regressor algorithms. As mentioned previously, prior exploration of long-form text generation has been limited. While \citet{mohri2024language} evaluate one variant---split conformal (SC)---on a small set of entities, we are not aware of prior work that has considered other uncertainty quantification methods in this setting.

\subsection{Iterative ``patching'' algorithms}

The first category of algorithms can be characterized as patching algorithms. Given a base method for calibration or conformal prediction, one iterates through groups $g \in \mathcal{G}$ in which the method does poorly on. At each iteration, the algorithm corrects the bias (i.e., patches up the function) on just that subset of examples (i.e., $g(x) = 1$). Once some stopping condition is met,\footnote{In the standard formulation of iterative patching, the stopping criteria is set as a function over the number of bins so that one can prove guarantees about algorithm (see \citet{roth2022uncertain}). In practice, we found this stopping criteria to be too conservative, and so we instead run iterative patching on the calibration and test sets concurrently and use the calibration set to determine the stopping iteration (i.e., we enforce early stopping once we can no longer make improvements on the calibration set).} the final, "patched up" function satisfies multi-group guarantees.

\paragraph{Calibration.}

For calibration, we consider \textit{Histogram Binning} (HB) \citep{zadrozny2001obtaining}, presented in Algorithm \ref{alg:hb}. This method, takes some base scoring function $f$ and discretizes the output space to a set of $p$-th level sets $S_p(f)$, as defined in Section \ref{sec:calibration}. Given some target grid of values $p\in{[\frac{1}{m}]}$, we round $f$ to the closest value in the grid
\begin{equation*}
    f'(x) = \argmin_{p\in{[\frac{1}{m}]}} | f(x) - p |.
\end{equation*}
Algorithm \ref{alg:hb} then applies a constant correction\footnote{In Algorithms \ref{alg:hb} and \ref{alg:ighb}, we assume true data distribution is given, and therefore we can calculate $\Delta_{p,g}$. In practice (and our experiments) $\Delta_{p,g}$ is estimated using a calibration set.} for each level set $S_p(f)$ in the grid, based on the calibration error of the model $f'$.

In Algorithm \ref{alg:ighb}, we present the multi-group version of histogram binning, known as \textit{Iterative Grouped Histogram Binning} (IGHB) \citep{hebert2018multicalibration}. In this algorithm, we instead apply a constant correction conditioned on $S_{p, g}$ (i.e., both the level set and group membership). At each step $t$, IGHB identifies $S_{p, g}$ for which the calibration error (weighted by the group size) is highest and then corrects it for this level set \textit{and} group. The algorithm then continues until some stopping condition is met, iteratively patching $f'$ for various groups $g \in \mathcal{G}$.

\begin{algorithm}[!t]
\caption{Histogram Binning (HB)}
\label{alg:hb}
\begin{algorithmic}[1]
\STATE \textbf{Input:} scoring function $f'$
\FOR{$p \in [\frac{1}{m}]$}
{
\STATE Set 
\[
    \hat{f}(x) = 
    \begin{cases} 
        f'(x) + \Delta_p(f') & \text{if } x \in S_p(f') \\
        f'(x) & \text{otherwise}
    \end{cases}
\]
}
\ENDFOR
\STATE \textbf{Output:} $\hat{f}$
\end{algorithmic}
\end{algorithm}
\begin{algorithm}[!htb]
\caption{Iterative Grouped Histogram Binning (IGHB)}
\label{alg:ighb}
\begin{algorithmic}[1]
\STATE \textbf{Input:} scoring function $f'$, max iterations $T$
\STATE Let $H_t(p, g) = P_D ( S_{p,g}(f_t) ) \Delta^2_{p,g}(f_t)$
\STATE Initialize $f_0 = f'$
\FOR{$t \in \{0, 1, \ldots, T-1 \}$}
    \STATE Set
    \[
        (p_t, g_t) = \argmax_{p \in \left[ \frac{1}{m} \right], g \in \mathcal{G}} H_t(p, g)
    \]
    \STATE Let $\Delta_t = \Delta_{p_t, g_t}$ and $S_t = S_{p_t, g_t}$
    \STATE Set 
    \[
        h_{t+1}(x) = 
        \begin{cases} 
        f_t(x) + \Delta_t(f_t) & \text{if } x \in S_t(f_t) \\
        f_t(x) & \text{otherwise}
        \end{cases}
    \]
    \STATE Set $f_{t+1} = h_{t+1}$
    \IF{$H_t(p_t, g_t) \ge H_{t-1}(p_{t-1}, g_{t-1})$} 
        \STATE Set t = t-1
        \BREAK
    \ENDIF
\ENDFOR
\STATE \textbf{Output:} $H_t$
\end{algorithmic}
\end{algorithm}

\paragraph{Conformal prediction.} We first present the \textit{Split Conformal} (SC) method \citep{shafer2008tutorial, gupta2022nested}. In particular, we consider the standard approach where one constructs a set of nested sets and each output set contains some subset $\mathcal{F(X)}_{t}$ of claims generated by the LLM.

Following \citet{mohri2024language}, we define these nested sets $\mathcal{T}$ as thresholds sets where each set $\mathcal{F}(L(X))$ contains the set of all individual claims $\{ x \in L(X) \mid f(x) > t \}$ for some scoring function $f$. More formally, we have that $\mathcal{F}(L(X))_{t \in \mathcal{T}}$ satisfies the nested sequence property if for $t, t' \in \mathcal{T}, t \le t'$, we have that $\mathcal{F}_t(L(X)) \subseteq \mathcal{F}_{t'}(L(X))$.

To construct these threshold sets, we have that
\begin{equation*}
    r(X,Y) = \inf \{ t \in \mathcal{T}, Y \in \mathcal{F}_t(L(X)) \}
\end{equation*}
where $r$ defines the minimum safe threshold such that $Y \in \mathcal{F}_t(L(X))$ for all $t > r(X,Y)$. Practically speaking, given some set of uncertainty scores $f(x)$ for each claim $x \in L(X)$, $r(X,Y)$ defines the minimum value such that any set of claims $\mathcal{F}_t(L(X)) = \{ x \in L(X) \mid f(x) \ge t \}$ will be entirely true if and only if $t \ge r(X,Y)$.

Given some calibration set $\hat{D}$ of size $n$ and some target error rate $\alpha$ (or target coverage $1-\alpha$), split conformal simply outputs the set $\mathcal{F}_{q_\alpha}(L(X))$ for any $X$, where $q_\alpha$ is the $\frac{\lceil (n+1)(1-\alpha) \rceil}{n}$th-quantile of scores $\{r(X_i, Y_i)\}_{i=1}^n$ for $X_i,Y_i \in \hat{D}$.

In Algorithm \ref{alg:mvsc}, we present the \textit{multivalid split conformal} (MVSC) prediction technique that closely resembles methods originally proposed in \citet{jung2022batch}. Similar to IGHB, we start with some base threshold (i.e., the threshold $q_\alpha$ obtained from using split conformal). Then at each iteration $t$, we find the group $g_t$ that has the worst squared coverage error $\Delta_{t,g}$, weighted by the size of the group $P(g_t(X) = 1)$. Then, we simply "patch" the thresholds for examples $\{ (X,Y) \mid g_t(X) = 1\}$, again using the $\frac{\lceil (n+1)(1-\alpha) \rceil}{n}$th-quantile of scores for $(X,Y)$ belong to group $g_t$. Like in IGHB, we continue patching the set of thresholds until some stopping criterion is met.

\begin{algorithm}[!t]
\caption{Multivalid Split Conformal (MVSC)}
\label{alg:mvsc}
\begin{algorithmic}[1]
\STATE \textbf{Input:} calibration set $\hat{D}$, LLM $L$, fact-level scoring function $f$, target error rate $\alpha$, split conformal threshold $q_\alpha$, max iterations $T$
\STATE Let $\mathcal{F}_{h_t}(L(X)) = \{ x \in L(X) \mid f(x) \ge h_t(X) \}$
\STATE Let $\Delta_{t, g} = P_{\mathcal{D}} (Y \in \mathcal{F}_{h_t}(L(X)) \mid g(X) = 1)$
\STATE Let $H_t(g) = P_{\mathcal{D}}(g(X) = 1) [(1 - \alpha) - \Delta_{t, g}]^2$

\STATE Initialize $h_0(X) = q_\alpha$

\FOR{$t \in \{0, 1, \ldots, T-1 \}$}
    \STATE Set
    \[
        g_t = \argmax_{g \in \mathcal{G}} H_t(g)
    \]
    \STATE Let $\hat{D}_t = \{ (X,Y) \in \hat{D} \mid g_t(X) = 1\}$
    \STATE Set $q_t$ to be the $\frac{\lceil (n+1)(1-\alpha) \rceil}{n}$th-quantile of scores $\{r(X_i, Y_i)\}$ for $X_i,Y_i \in \hat{D}_t$
    \STATE Set 
    \[
        h_{t+1}(X) = 
        \begin{cases} 
        q_t & \text{if } g_t(X) = 1 \\
        f_t(X) & \text{otherwise}
        \end{cases}
    \]
    \IF{$H_t(g_t) \ge H_{t-1}(g_{t-1})$} 
        \STATE Set t = t-1
        \BREAK
    \ENDIF
\ENDFOR
\STATE \textbf{Output:} $h_t$
\end{algorithmic}
\end{algorithm}

\subsection{Linear regressor algorithms}

Next, we consider algorithms that instead solve an optimization problem for the purpose of calibration and conformal prediction. In these cases, one can naturally make them multi-group/valid by including group-membership (i.e., $g(X) = 1$ for all $g \in \mathcal{G}$) in the optimization problem itself. Formally, we describe these linear regression based methods in Algorithms \ref{alg:linear_regressor} and \ref{alg:group_regressor}. Presented in this way, the methods for calibration vs. conformal prediction is reduced to a choice of loss function $L$. Again, we assume one has access to some calibration set for which one solves the optimization problem on.

\paragraph{Calibration.} For calibration, one can choose $L$ to be binary cross-entropy loss. In doing so, Algorithm \ref{alg:linear_regressor} then describes \textit{Platt Scaling} (PS) \citep{platt1999probabilistic}, which can be described as fitting a logistic regression model to some set of model outputs to obtain calibrated probability scores.\footnote{A related calibration method to Platt scaling (PS) is temperature scaling (TS) \citep{guo2017calibration}, which was originally introduced for calibrating neural networks for multiclass classification and has been incorporated in work on calibrating NLP models \citep{sicilia2024deal}. We note, however, that in the binary classification setting (e.g., our setting where we identify if an output is correct or not), TS is mathematically equivalent to PS when there is no bias term and the weight takes on the form $\frac{1}{\tau}$, where $\tau$ is the temperature learned in TS.}
Algorithm \ref{alg:group_regressor} describes the multi-calibrated version of Platt Scaling. While not explicitly derived in their work, this multicalibration formulation can be traced back to \citet{gopalan2022low}, who establish a hierarchy of notions for multicalibration and analyze multicalibration on functions trained with linear loss. Going forward, we refer to this method as \textit{Group Conditional Unbiased Logistic Regression} (GCULR).

\paragraph{Conformal prediction.} For conformal prediction, we instead choose $L$ to be pinball loss. We refer to the non-group version of this method (Algorithm \ref{alg:linear_regressor}) as \textit{Conformalized Quantile Regression} (CQR) \citep{romano2019conformalized}, in which given some target coverage $1 - \alpha$, we fit a linear quantile regression model that minimizes pinball loss. 

In our conformal prediction setting, as described in Section \ref{sec:conformal}, $X$ is an entire biography, or set of independent claims. Thus, to adapt quantile regression to long-form generation, we propose setting $f(X)$ to be a vector of uncertainty scores for each claim $x \in X$. Like in split conformal, the target is then the minimum threshold $r(X,Y)$ for which all claims above it are correct. In the multivalid case, we then add group features $g(X)$ to the optimization problem. A version of Algorithm $\ref{alg:group_regressor}$ was first presented by \citet{jung2022batch}, and going forward, we will refer to this method as \textit{Group Conditional Conformalized Quantile Regression} (GCCQR).

We note that in our experiments, each biography generated by the LLM may have a different number of claims, a setting in which prior work on conformal quantile regression does not account for. Consequently, we propose using interpolation to (un)squeeze the set of scores to a vector $f(X)$ of fixed size ($K=25$ in our experiments). While \citet{mohri2024language} only show that split conformal can be applied to this type of setting, our experiments demonstrate that quantile regression methods achieve similar performance for marginal (CQR) and multigroup (GCCQR) methods (Section \ref{sec:results}).

\begin{algorithm}[!t]
\caption{Linear Regressor}
\label{alg:linear_regressor}
\begin{algorithmic}[1]
\STATE \textbf{Input:} data distribution $\mathcal{D}$, scoring function $f$, loss function $L$
\STATE Set 
\[
    \hat{\lambda} = 
    \argmin_{\lambda}
    \mathbb{E}_{(X,Y) \sim \mathcal{D}} \left[ L\left( f(X; \lambda), Y \right) \right]
\]
s.t. $f(X; \lambda) = \lambda_0 + \lambda_1 f(X)$
\STATE \textbf{Output:}$f(X; \hat{\lambda})$
\end{algorithmic}
\end{algorithm}

\begin{algorithm}[!t]
\caption{Group-conditional Linear Regressor}
\label{alg:group_regressor}
\begin{algorithmic}[1]
\STATE \textbf{Input:} data distribution $\mathcal{D}$, scoring function $f$, loss function $L$, set of groups $\mathcal{G}$
\STATE Set 
\[
    \hat{\lambda} = 
    \argmin_{\lambda}
    \mathbb{E}_{(X,Y) \sim \mathcal{D}} \left[ L\left( f(X;\lambda), Y \right) \right]
\]
s.t. $f(X; \lambda) = \lambda_0 + \lambda_1 f(X) + \sum_{\lambda_g \in \mathcal{G}} \lambda_g g(x)$
\STATE \textbf{Output:}$f(X; \hat{\lambda})$
\end{algorithmic}
\end{algorithm}

\begin{table*}[hbt!]
\centering
\begin{tabular}{c c c | c | c c | c c }
\toprule
\textbf{Model} & \textbf{Base Score} & \textbf{Metric} & \textbf{Uncalibrated} & \textbf{HB} & \textbf{IGHB} & \textbf{PS} & \textbf{GCULR} \\
\toprule
\multirow{9}{*}{Llama} 
& \multirow{3}{*}{self-consistency} 
& marginal     & 0.32291     & 0.00875   & \textbf{0.00038} & 0.00022 & \textbf{0.00015}* \\
& & group max  & 0.42343     & 0.07711   & \textbf{0.01481} & 0.05791 & \textbf{0.00628}* \\ 
& & group mean & 0.33352     & 0.01597   & \textbf{0.00289} & 0.00636 & \textbf{0.00111}* \\ 
\cmidrule{2-8}
& \multirow{3}{*}{P(True)} 
& marginal     & 0.11768     & 0.00451   & \textbf{0.00021}* & 0.00036 & \textbf{0.00022} \\
& & group max  & 0.20701     & 0.06988   & \textbf{0.01798} & 0.06025 & \textbf{0.00697}* \\ 
& & group mean & 0.12682     & 0.01176   & \textbf{0.00328} & 0.00654 & \textbf{0.00145}* \\ 
\cmidrule{2-8}
& \multirow{3}{*}{verb. conf.} 
& marginal     & 0.01642     & \textbf{0.00014}   & 0.00055 & \textbf{0.00013}* & 0.00023 \\
& & group max  & 0.06645     & 0.06634   & \textbf{0.01315} & 0.06709 & \textbf{0.00730}* \\ 
& & group mean & 0.02447     & 0.00738   & \textbf{0.00357} & 0.00750 & \textbf{0.00154}* \\ 
\midrule
\multirow{9}{*}{Mistral} 
& \multirow{3}{*}{self-consistency} 
& marginal     & 0.30706     & 0.01163   & \textbf{0.00026} & 0.00029 & \textbf{0.00013}* \\
& & group max  & 0.43659     & 0.08372   & \textbf{0.01660} & 0.05729 & \textbf{0.00487}* \\ 
& & group mean & 0.32067     & 0.01988   & \textbf{0.00269} & 0.00726 & \textbf{0.00097}* \\ 
\cmidrule{2-8}
& \multirow{3}{*}{P(True)} 
& marginal     & 0.06291     & 0.00074   & \textbf{0.00031} & 0.00047 & \textbf{0.00015}* \\
& & group max  & 0.12293     & 0.06942   & \textbf{0.01417} & 0.07054 & \textbf{0.00587}* \\ 
& & group mean & 0.07173     & 0.00896   & \textbf{0.00309} & 0.00886 & \textbf{0.00132}* \\ 
\cmidrule{2-8}
& \multirow{3}{*}{verb. conf.} 
& marginal     & 0.22229     & 0.00047   & \textbf{0.00036} & 0.00034 & \textbf{0.00015}* \\
& & group max  & 0.33763     & 0.06922   & \textbf{0.01453} & 0.07080 & \textbf{0.00653}* \\ 
& & group mean & 0.23230     & 0.00869   & \textbf{0.00315} & 0.00878 & \textbf{0.00128}* \\ 
\bottomrule
\end{tabular}
\caption{We generate biographies using Llama 2 7B Chat and Mistral 7B Instruct for entities from \textsc{Bio-NQ} and compare each calibration method (HB, PS) against its multicalibration counterpart (IGHB, GCULR) on \textbf{ASCE}, \textbf{max gASCE}, and \textbf{average gASCE} ($\downarrow$ better). We test each method using the base scores: self-consistency, P(True), and verbalized confidence. We bold the better-performing method for each pairing and use * to denote the best-performing method across all methods.}
\label{tab:multicalibration_asce_nq}
\end{table*}


\begin{table*}[htb!]
\centering
\begin{tabular}{c c c | c | c c | c c }
\toprule
\textbf{Model} & \textbf{Base Score} & \textbf{Metric} & \textbf{Uncalibrated} & \textbf{HB} & \textbf{IGHB} & \textbf{PS} & \textbf{GCULR} \\
\toprule
\multirow{9}{*}{Llama} 
& \multirow{3}{*}{self-consistency} 
& marginal     & 0.475     & 0.169   & \textbf{0.148} & 0.152 & \textbf{0.143}* \\
& & group max  & 0.535     & 0.323   & \textbf{0.247} & 0.285 & \textbf{0.235}* \\
& & group mean & 0.479     & 0.169   & \textbf{0.148} & 0.152 & \textbf{0.143}* \\
\cmidrule{2-8}
& \multirow{3}{*}{P(True)} 
& marginal     & 0.274     & 0.165   & \textbf{0.152} & 0.157 & \textbf{0.149}* \\
& & group max  & 0.341     & 0.315   & \textbf{0.261} & 0.305 & \textbf{0.250}* \\
& & group mean & 0.277     & 0.165   & \textbf{0.152} & 0.157 & \textbf{0.148}* \\
\cmidrule{2-8}
& \multirow{3}{*}{verb. conf.} 
& marginal     & 0.177     & 0.161   & \textbf{0.152} & 0.161 & \textbf{0.150}* \\
& & group max  & 0.270     & 0.311   & \textbf{0.253} & 0.311 & \textbf{0.248}* \\
& & group mean & 0.177     & 0.160   & \textbf{0.152} & 0.160 & \textbf{0.149}* \\
\midrule
%
\multirow{9}{*}{Mistral} 
& \multirow{3}{*}{self-consistency} 
& marginal     & 0.471     & 0.186   & \textbf{0.164} & 0.159 & \textbf{0.152}* \\ 
& & group max  & 0.554     & 0.333   & \textbf{0.285} & 0.250 & \textbf{0.235}* \\ 
& & group mean & 0.477     & 0.186   & \textbf{0.164} & 0.158 & \textbf{0.152}* \\ 
\cmidrule{2-8}
& \multirow{3}{*}{P(True)} 
& marginal     & 0.237     & 0.175   & \textbf{0.164} & 0.174 & \textbf{0.161}* \\
& & group max  & 0.304     & 0.318   & \textbf{0.259} & 0.317 & \textbf{0.249}* \\
& & group mean & 0.237     & 0.175   & \textbf{0.164} & 0.174 & \textbf{0.160}* \\
\cmidrule{2-8}
& \multirow{3}{*}{verb. conf.} 
& marginal     & 0.397     & 0.175   & \textbf{0.164} & 0.175 & \textbf{0.161}* \\
& & group max  & 0.427     & 0.318   & \textbf{0.259} & 0.318 & \textbf{0.249}* \\
& & group mean & 0.398     & 0.175   & \textbf{0.164} & 0.174 & \textbf{0.160}* \\
\bottomrule
\end{tabular}
\caption{We generate biographies using Llama 2 7B Chat and Mistral 7B Instruct for entities from \textsc{Bio-NQ} and compare each calibration method (HB, PS) against its multicalibration counterpart (IGHB, GCULR) on \textbf{Brier score} ($\downarrow$ better) \textbf{marginally} across the entire dataset, as well as within each subgroup (in terms of \textbf{max} and \textbf{mean} over all groups). We test each method using the base scores: self-consistency, P(True), and verbalized confidence. We bold the better-performing method for each pairing and use * to denote the best-performing method across all methods.}
\label{tab:multicalibration_brier_nq}
\end{table*}

\section{Empirical evaluation}

We focus our empirical evaluation on the problem of biography generation, which we contend serves as a very suitable testbed for evaluating factuality and has been used as a benchmark in a variety of works in recent years. Outputting biographies offers one the ability to evaluate not only a set of objective and specific claims but also on a wide range of topics, which in turn allows us to explore a rich set of group functions for each person. Moreover, bias within biography generation has long been a studied issue, further motivating the problem of ensuring group-conditional guarantees. 
Like \citet{min2023factscore}, we use a language model to automate the process of decomposing biographies into claims and evaluating for factuality (Appendix \ref{appx:generating}).

\paragraph{Dataset}

We evaluate on a large set of biographies by extracting 8,541 entities from the Natural Questions dataset \citep{kwiatkowski2019natural}, which consists of real queries issued to the Google search engine. We denote this dataset as \textsc{Bio-NQ}. Our motivation for choosing Natural Questions is that these extracted human entities should serve as a representative sample of public figures that users may prompt an LLM about. For each question, we select all entities in either the question's short answer or accompanying Wikipedia article. We then attempt to match them to their corresponding Wikidata entry. If a match exists and its Wikidata page's property, \textit{if instance of}, is equal to the value, \textit{human}, we add the entity to our dataset \textsc{Bio-NQ}.

\paragraph{Collecting group features}

To obtain groups for each person found in our dataset, we extract properties by scraping Wikidata for each entity and identifying ones that are commonly shared among entities in \textsc{Bio-NQ}. The exact group attributes we use in our experiments are described in Appendix \ref{appx:additional_exp_details}. To form groups $\mathcal{G}$ from these attributes, we take all $1$ and $2$-way combinations of attributes and the values they take on, giving us $|\mathcal{G}| = 77$ subgroups.

\paragraph{Generating confidence scores}

The algorithms described in Section \ref{sec:methods} require a base scoring function. For experiments, we use the following:

\begin{enumerate}
    \item \textbf{Self-consistency} \citep{wang2022self}: Our first score is a frequency-based scoring function inspired by \textit{self-consistency}. To score each claim found in a generated biography, we prompt the LLM to output a biography $M$ additional times. We use the proportion of times the claim is contained in the additional reference generations as the uncertainty score. We automate the calculation of this score using BM25 and AlignScore \citep{zha2023alignscore} (See Appendix \ref{appx:base_scoring_fns}).

    \item \textbf{P(True)} \citep{kadavath2022language}: For each biographical claim, we prompt the LLM to assess whether it is true or false. We then output the ratio of next token probabilities of the tokens for ``true'' and ``false'': $\frac{P(True)}{P(True) + P(False)}$. 

    \item \textbf{Verbalized confidence} \citep{tian2023just}: To output \textit{verbalized confidence} as an uncertainty estimate, one prompts the LLM to directly output their confidence level in its response. We originally tried having the model rate its confidence numerically (e.g., output an integer between 1-5, 1-10, 1-100, etc.). However, we found these base scores to be somewhat unreliable. Instead, we ask the LLM to rate its confidence in each individual claim using integers between 1 and 5. We then output a weighted sum of the next-token probabilities for the tokens ``1'' through ``5'': $\sum_{r=1}^5 r \times P(r)$.
\end{enumerate}

\section{Empirical Results}\label{sec:results}

\begin{figure*}[tb!]
    \centering
    \begin{subfigure}[t]{\columnwidth}
        \centering
        \includegraphics[width=1.0\textwidth]{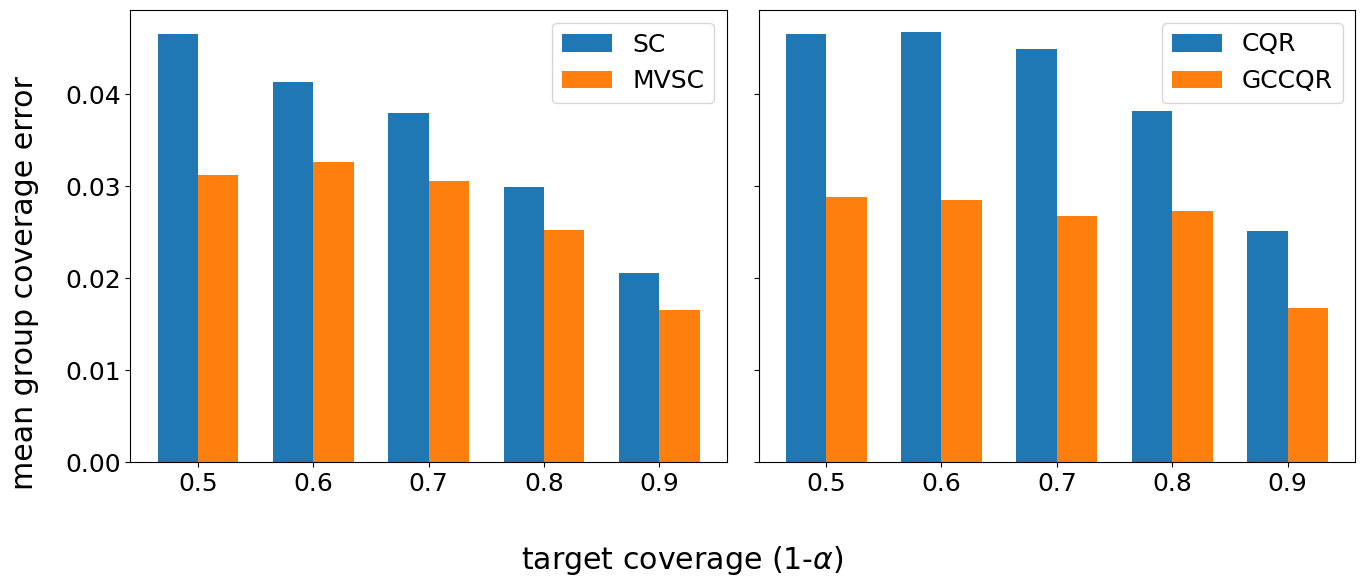}
        \caption{LLama 2 7B Chat; self-consistency}
        \label{fig:conformal_nq_llama_selfconsistency}
    \end{subfigure}
    \begin{subfigure}[t]{\columnwidth}
        \centering
        \includegraphics[width=1.0\textwidth]{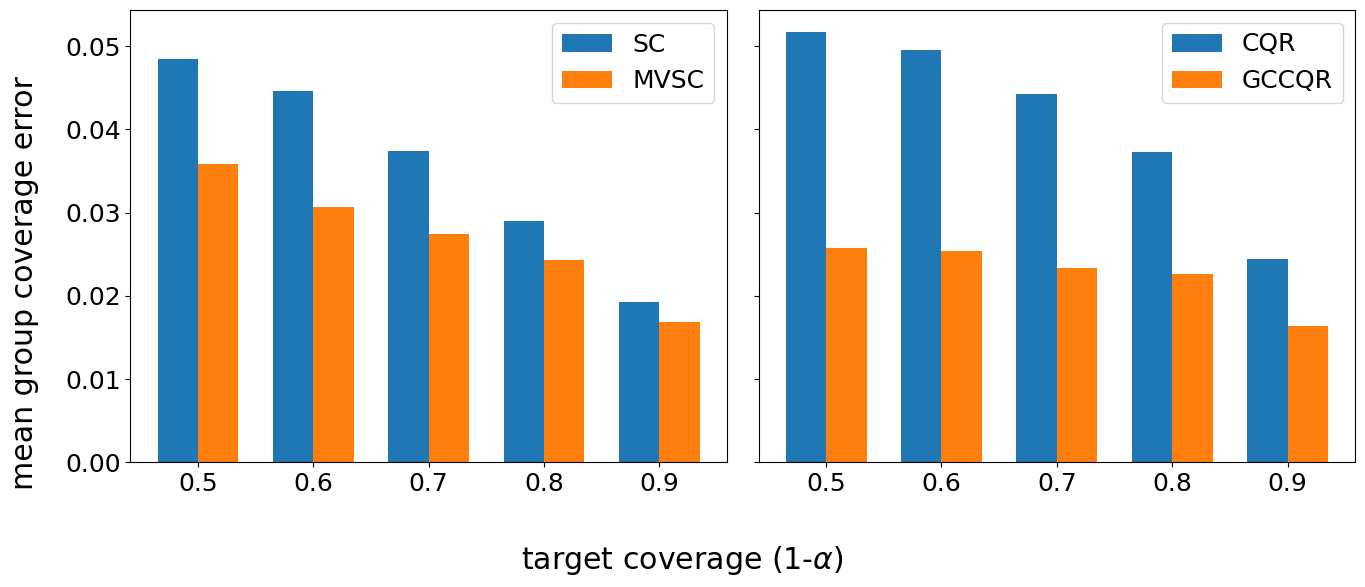}
        \caption{Mistral 7B Instruct; self-consistency}
        \label{fig:conformal_nq_mistral_selfconsistency}
    \end{subfigure}
    \begin{subfigure}[t]{\columnwidth}
        \centering
        \includegraphics[width=1.0\textwidth]{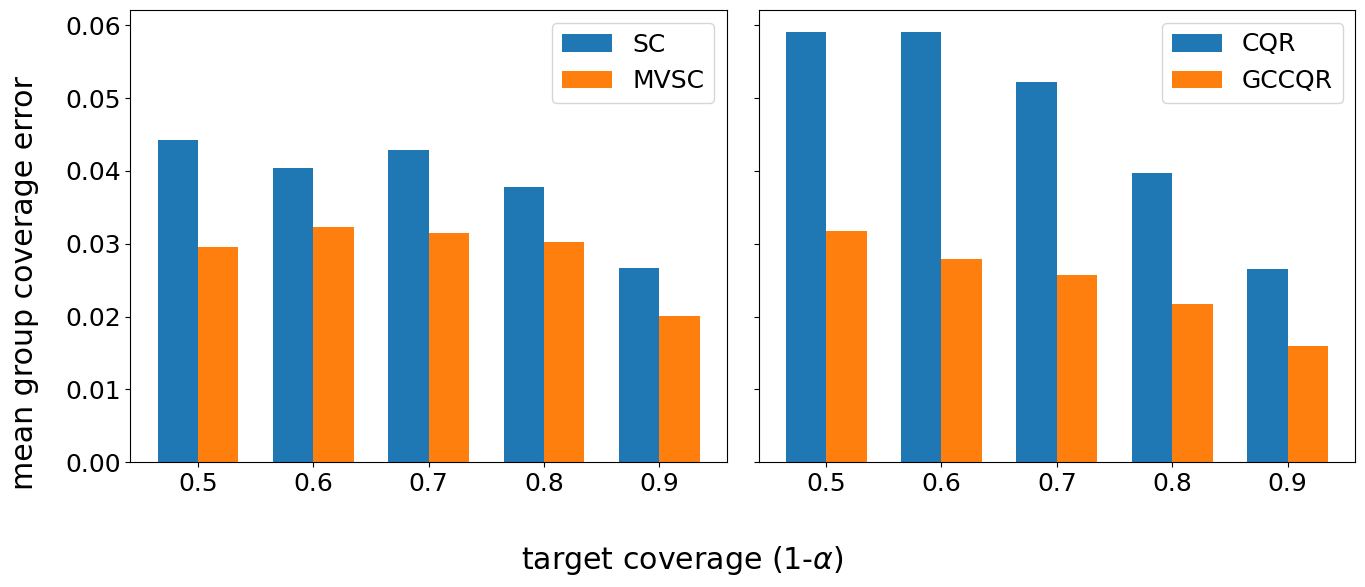}
        \caption{LLama 2 7B Chat; P(True)}
        \label{fig:conformal_nq_llama_ptrue}
    \end{subfigure}
    \begin{subfigure}[t]{\columnwidth}
        \centering
        \includegraphics[width=1.0\textwidth]{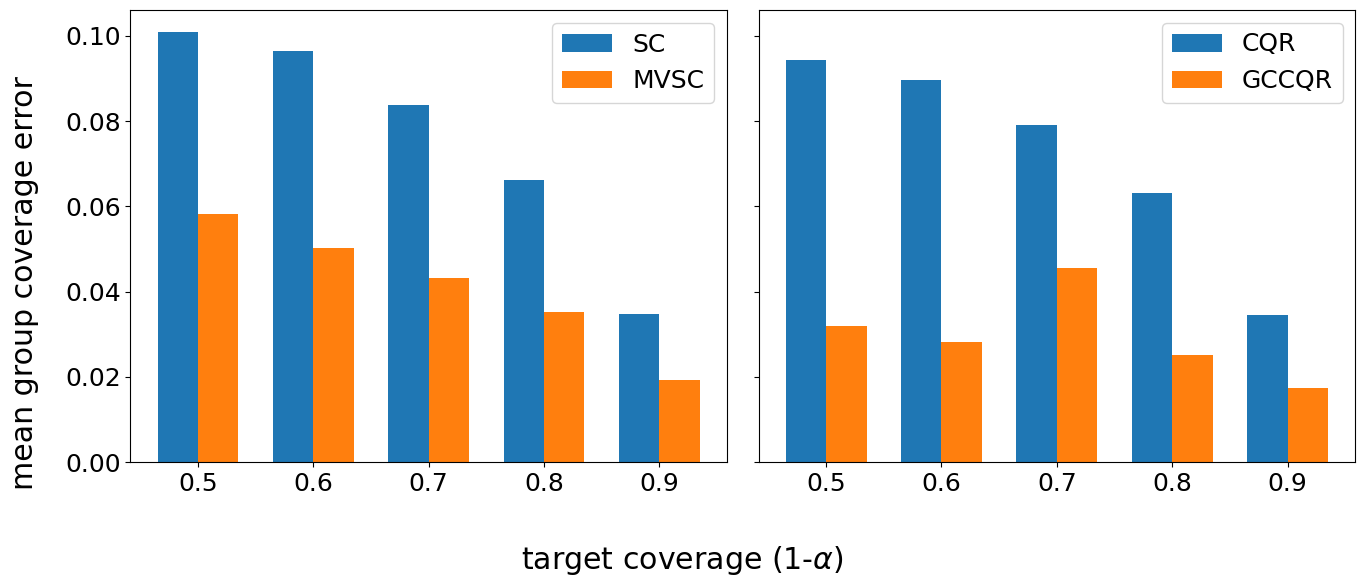}
        \caption{Mistral 7B Instruct; P(True)}
        \label{fig:conformal_nq_mistral_ptrue}
    \end{subfigure}
    \begin{subfigure}[t]{\columnwidth}
        \centering
        \includegraphics[width=1.0\textwidth]{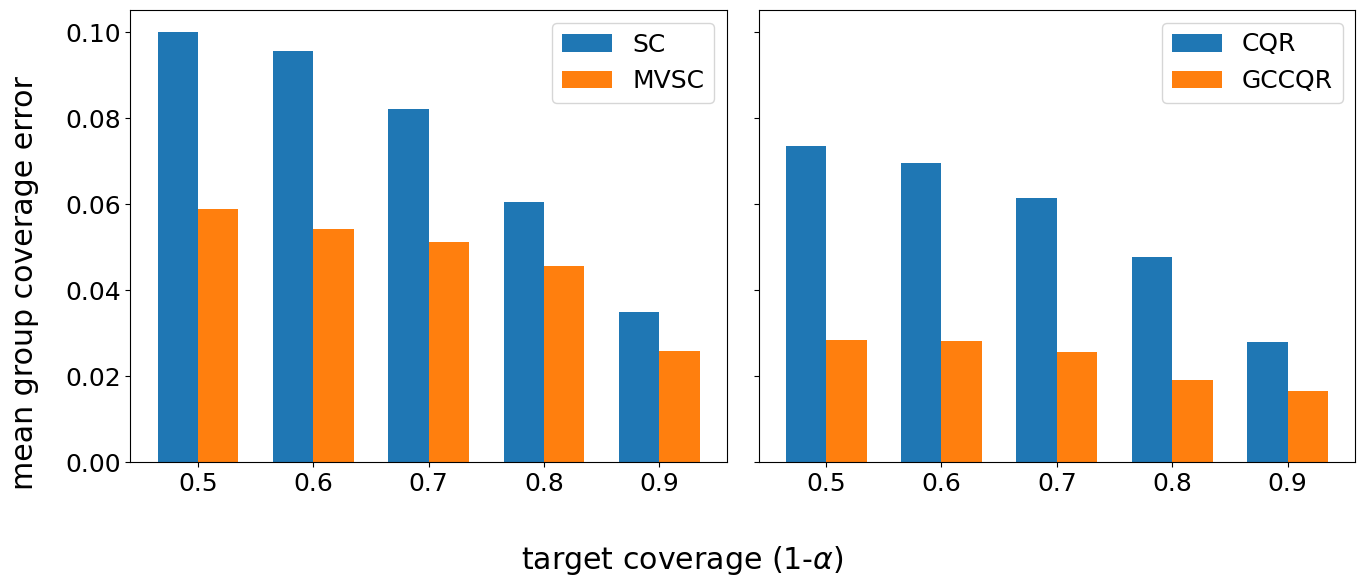}
        \caption{LLama 2 7B Chat; verbalized confidence}
        \label{fig:conformal_nq_llama_verbconf}
    \end{subfigure}
    \begin{subfigure}[t]{\columnwidth}
        \centering
        \includegraphics[width=1.0\textwidth]{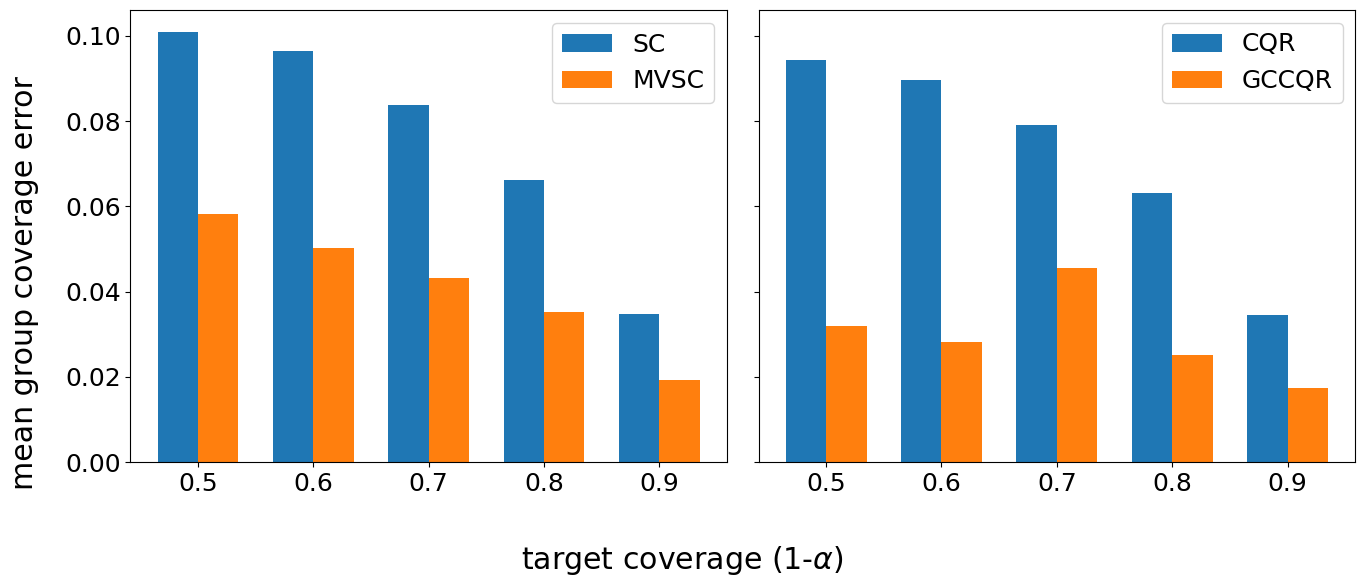}
        \caption{Mistral 7B Instruct; verbalized confidence}
        \label{fig:conformal_nq_mistral_verbconf}
    \end{subfigure}
    \caption{For each target coverage, we run conformal methods (\textcolor{mplblue}{\textbf{blue}}: SC, CQR) and their multigroup counterparts (\textcolor{mplorange}{\textbf{orange}}: MVSC, GCCQR) on \textsc{Bio-NQ} using the following base uncertainty scoring functions: \textbf{(a, b)} self-consistency, \textbf{(c, d)} P(True), and \textbf{(e, f)} verbalized confidence. We evaluate on generations from \textbf{(a, c, e)} Llama 2 7B Chat and \textbf{(b, d, f)} Mistral 7B Instruct. We calculate the average coverage error across all groups and plot them side by side for each pairing.}
    \label{fig:conformal_nq}
\end{figure*}

To assess the efficacy of the methods introduced in Section~\ref{sec:methods}, we present results for the task of biography generation using outputs from Llama 2 7B Chat \citep{touvron2023llama} and Mistral 7B Instruct v0.2 \citep{jiang2023mistral}. We randomly split the entities into 80-20 calibration-test splits, averaging results over 10 randomly generated splits.

We stress that the primary goal (and novelty) of our work is to evaluate---as it pertains to metrics for uncertainty quantification---the (1) efficacy and failures of marginal methods  and (2) the extent to which multi-group methods improve over them. In supplementary results found in Appendix \ref{appx:analysis}, we include additional analysis comparing how marginal and group-conditional methods behave. Tables  \ref{tab:examples_llama2_retain_more}, \ref{tab:examples_mistral_retain_more}, \ref{tab:examples_llama2_nonempty}, and \ref{tab:examples_mistral_nonempty} of the appendix provide examples of the types of outputs produced by both marginal and multivalid conformal methods.

\paragraph{Calibration.} 


 In Table \ref{tab:multicalibration_asce_nq}, we report ASCE, max gASCE, and mean gASCE, comparing each calibration method (HB, PS) against its multicalibration counterpart (IGHB, GCULR) across various base scoring functions. We find that marginal calibration methods (HB, PS) are able to correct the uncalibrated uncertainty scores, significantly decreasing ASCE. However, when examining max and mean gASCE, we find that these methods do not ensure strong guarantees for uncertainty when evaluating within different subgroups. The discrepancy is particularity large in cases where the marginal method performs well w.r.t. marginal ASCE. For example, when applying PS to self-consistency scores and comparing ASCE to max group gASCE, we observe that there exists some subgroup for which the calibration error is approximately \textbf{263x} and \textbf{198x} (for Llama and Mistral output respectively) worse on that particular subgroup compared to the dataset as a whole.
 
In contrast, the multicalibration variants of both the patching (IGHB) and linear regression (GCULR) techniques significantly outperform HB and PS in terms of max and mean gASCE across all experimental settings. Our results provide strong evidence that regardless of the model, base scoring function, or algorithm type, incorporating information about the subgroups in some meaningful way will substantially correct biases that marginal methods exhibit.

Even more surprising is that when considering just (marginal) ASCE across the entire dataset, incorporating group features improves performance as well. Aside from the experiments calibrating verbalized confidence scores on Llama 2 7B Chat generations, where HB and PS perform particularly well, the multicalibration variant outperforms the marginal method every time, with GCULR being the best method in almost all cases. While improving marginal calibration is not the primary focus of our work, these results suggest that even if one does not specifically require parity for specific subgroups, collecting additional group features and applying multicalibration (as opposed to vanilla calibration) can still be extremely beneficial for generating better-calibrated uncertainty scores. 

Finally, to evaluate fact-level uncertainty more holistically, we also consider the Brier score, which is the mean squared error between the uncertainty score function $f(X)$ and the true label $Y$. While not a direct measure of (multi)calibration like ASCE, the Brier score is still useful in certain settings for quantifying the efficacy of the algorithms we consider, quantifying desirable properties of calibration that are not captured by calibration error \citep{brocker2009reliability, liu2025calibrating}. In Table \ref{tab:multicalibration_brier_nq}, we report the Brier score, also both marginally and across groups. Similar to our analysis of ASCE, we again see that standard calibration methods exhibit failures when evaluating at the group level. However, IGHB and GCULR outperform HB and PS respectively across all metrics.

\paragraph{Conformal prediction.} For the problem of uncertainty at the biography level, we apply the vanilla conformal prediction methods SC and CQR and their multivalid counterparts, MVSC and GCCQR.\footnote{To compare methods qualitatively, we provide illustrative example outputs in Appendix \ref{appx:analysis}, Tables \ref{tab:examples_llama2_retain_more}, \ref{tab:examples_mistral_retain_more}, \ref{tab:examples_llama2_nonempty}, and \ref{tab:examples_mistral_nonempty}.}
We choose target coverages of between $0.5$ to $0.9$, evaluating on biographies generated by Llama 2 7B Chat and Mistral 7B Instruct.\footnote{Although we evaluate on a wide set of target coverages $1-\alpha$, conformal prediction makes more sense only for higher target coverages (e.g., $0.8$ or higher), since lower coverage guarantees can often be too weak to be useful in practice.}

We corroborate \citet{mohri2024language}'s findings that (standard) conformal prediction methods are able to achieve close to perfect coverage on biography generation. Specifically, we show that both SC and CQR achieve target coverages (Appendix \ref{appx:analysis}, Figure \ref{fig:conformal_nq_additional}). Moreover, there is little difference between the two in terms of the average number of abstentions and facts per biography retained. 
However, when evaluating coverage across individual subgroups, we find that both methods have some level of error. In Figure \ref{fig:conformal_nq}, we compare the mean absolute coverage error across all subgroups for each target coverage and find that SC and CQR exhibit high mean errors (of up to $0.1$ in some cases), despite achieving almost no error when evaluated (marginally) across the entire dataset (Figure \ref{fig:conformal_nq_additional}). 

Again, we investigate whether incorporating subgroup information can correct these biases.
Here, the message is clear---multivalid conformal methods improve coverage error at the group level, regardless of the model, base scoring function, or algorithm type (Figure \ref{fig:conformal_nq}). We note however that we do not observe the same performance gains as found for calibration (Table \ref{tab:multicalibration_asce_nq}), where group-conditional methods sometimes outperform marginal ones by an entire order of magnitude. This finding may result in part due to the smaller calibration set or the possibility that (multivalid) conformal prediction for LLMs is a more challenging problem. We leave further investigation of this observation to future work.



\section{Conclusion}

In this paper, we conduct an extensive study on uncertainty quantification for long-form text generation. We focus on two forms of uncertainty---claim-level (calibration) and biography-level (conformal prediction)---and present a variety of methods for these settings. 
We empirically validate that marginal methods for calibration and conformal prediction perform well when evaluated across the entire dataset. However, when looking at subgroup performance, we find that performance consistently degrades.
Introducing two categories of algorithms (iterative patching and linear regression), we demonstrate that by accounting for additional groups, multicalibration and multivalid conformal prediction methods correct the aforementioned biases of marginal-guarantee counterparts. We consider these empirical results to establish a benchmark for this setting and hope that our findings will motivate future work in this area.

\bibliography{docs/main}

\newpage

\appendix

\onecolumn

\title{Multi-group Uncertainty Quantification for Long-form Text Generation\\(Supplementary Material)}
\maketitle

\section{Additional analysis and results}\label{appx:analysis}

We provide additional analysis of the methods studied in our paper. For conciseness, we conduct this analysis on methods applied to \textit{self-consistency} scores only. We note, however, that similar findings can be made when applying such methods to P(True) or verbalized confidence.

\paragraph{Calibration.}

\begin{table*}[b!]
\centering
\renewcommand{\arraystretch}{1.1}
\begin{tabular}{l l c c c}
\toprule
& & \textbf{HB} & \textbf{IGHB} & $\Delta$ \\
\toprule
\multirow{5}{*}{Top 5 $\Delta$}
& \textbf{\# Wiki prop.} = \textit{Low} \textcolor{ForestGreen}{\&} \textbf{nationality} = \textit{EU/ME} & 0.0771 & 0.0148 & -0.0623 \\
& \textbf{\# Wiki prop.} = \textit{Low} \textcolor{ForestGreen}{\&} \textbf{has IMDb ID} = \textit{True} & 0.0540 & 0.0055 & -0.0485 \\
& \textbf{\# Wiki prop.} = \textit{Low} \textcolor{ForestGreen}{\&} \textbf{sport} = \textit{False} & 0.0429 & 0.0031 & -0.0398 \\
& \textbf{\# Wiki prop.} = \textit{Low} \textcolor{ForestGreen}{\&} \textbf{sex or gender} = \textit{female} & 0.0395 & 0.0033 & -0.0362 \\
& \textbf{\# Wiki prop.} = \textit{Low} \textcolor{ForestGreen}{\&} \textbf{nationality} = \textit{NA} & 0.0379 & 0.0042 & -0.0337 \\
\midrule
Min $\Delta$
& \textbf{\# Wiki prop.} = \textit{Medium} \textcolor{ForestGreen}{\&} \textbf{nationality} = \textit{APAC} & 0.0114 & 0.0088 & -0.0026 \\
\bottomrule
\end{tabular}
\renewcommand{\arraystretch}{1}
\caption{
[\textbf{Calibration} on \textit{self-consistency} scores] Using outputs from \textbf{Llama 2 7B Chat} on \textsc{Bio-NQ}, we calculate the ASCE for each group using \textbf{HB} and \textbf{IGHB} as well as the difference in ASCE ($\Delta$) between the two methods. First, we then present the top 5 groups according ($\Delta$) where \textbf{top} corresponds to groups for which the multicalibration method achieves the biggest improvement (most negative change $\Delta$). In our experiments, we find that IGHB improves over HB for all groups, and so as reference, we also present the group with the minimum amount of change between IGHB and HB.
}
\label{tab:asce_hb_best_worst_llama2}
\end{table*}


\begin{table*}[btp!]
\centering
\renewcommand{\arraystretch}{1.1}
\begin{tabular}{l l c c c}
\toprule
& & \textbf{HB} & \textbf{IGHB} & $\Delta$ \\
\toprule
\multirow{5}{*}{Top 5 $\Delta$}
& \textbf{\# Wiki prop.} = \textit{Low} \textcolor{ForestGreen}{\&} \textbf{nationality} = \textit{EU/ME} & 0.0837 & 0.0166 & -0.0671 \\
& \textbf{\# Wiki prop.} = \textit{Low} \textcolor{ForestGreen}{\&} \textbf{has IMDb ID} = \textit{True}  & 0.0718 & 0.0097 & -0.0621 \\
& \textbf{\# Wiki prop.} = \textit{Low} \textcolor{ForestGreen}{\&} \textbf{sport} = \textit{False} & 0.0505 & 0.0059 & -0.0446 \\
& \textbf{\# Wiki prop.} = \textit{Low} \textcolor{ForestGreen}{\&} \textbf{nationality} = \textit{NA} & 0.0458 & 0.0088 & -0.0370 \\
& \textbf{\# Wiki prop.} = \textit{Low} \textcolor{ForestGreen}{\&} \textbf{sex or gender} = \textit{female} & 0.0414 & 0.0045 & -0.0369 \\
\midrule
Min $\Delta$
& \textbf{\# Wiki prop.} = \textit{Medium} \textcolor{ForestGreen}{\&} \textbf{nationality} = \textit{APAC} & 0.0118 & 0.0098 & -0.0021 \\
\bottomrule
\end{tabular}
\renewcommand{\arraystretch}{1}
\caption{
[\textbf{Calibration} on \textit{self-consistency} scores] Using outputs from \textbf{Mistral 7B Instruct} on \textsc{Bio-NQ}, we calculate the ASCE for each group using \textbf{HB} and \textbf{IGHB} as well as the difference in ASCE ($\Delta$) between the two methods. First, we then present the top 5 groups according ($\Delta$) where \textbf{top} corresponds to groups for which the multicalibration method achieves the biggest improvement (most negative change $\Delta$). In our experiments, we find that IGHB improves over HB for all groups, and so as reference, we also present the group with the minimum amount of change between IGHB and HB.
}
\label{tab:asce_hb_best_worst_mistral}
\end{table*}


\begin{table*}[btp!]
\centering
\renewcommand{\arraystretch}{1.1}
\begin{tabular}{l l c c c}
\toprule
& & \textbf{PS} & \textbf{GCULR} & $\Delta$ \\
\toprule
\multirow{6}{*}{Top 5 $\Delta$}
& \textbf{\# Wiki prop.} = \textit{Low} \textcolor{ForestGreen}{\&} \textbf{nationality} = \textit{EU/ME} & 0.0579 & 0.0063 & -0.0516 \\
& \textbf{\# Wiki prop.} = \textit{Low} \textcolor{ForestGreen}{\&} \textbf{has IMDb ID} = \textit{True} & 0.0375 & 0.0030 & -0.0345 \\
& \textbf{\# Wiki prop.} = \textit{Low} \textcolor{ForestGreen}{\&} \textbf{sport} = \textit{False} & 0.0266 & 0.0013 & -0.0253 \\
& \textbf{\# Wiki prop.} = \textit{Low} \textcolor{ForestGreen}{\&} \textbf{sex or gender} = \textit{female} & 0.0243 & 0.0034 & -0.0209 \\
& \textbf{\# Wiki prop.} = \textit{Low} \textcolor{ForestGreen}{\&} \textbf{nationality} = \textit{NA} & 0.0338 & 0.0031 & -0.0306 \\
\cline{2-5}
& Mean & 0.0338 & 0.0031 & -0.0306 \\
\midrule
\multirow{6}{*}{Bottom 5 $\Delta$}
& \textbf{has IMDb ID} = \textit{True} \textcolor{ForestGreen}{\&} \textbf{nationality} = \textit{APAC} & 0.0015 & 0.0018 & 0.0003 \\
& \textbf{\# Wiki prop.} = \textit{Low} \textcolor{ForestGreen}{\&} \textbf{sport} = \textit{True} & 0.0028 & 0.0033 & 0.0005 \\
& \textbf{has IMDb ID} = \textit{False} \textcolor{ForestGreen}{\&} \textbf{nationality} = \textit{NA} & 0.0005 & 0.0012 & 0.0007 \\
& \textbf{has IMDb ID} = \textit{False} \textcolor{ForestGreen}{\&} \textbf{sex or gender} = \textit{female} & 0.0011 & 0.0019 & 0.0008 \\
& \textbf{nationality} = \textit{APAC} \textcolor{ForestGreen}{\&} \textbf{sex or gender} = \textit{female} & 0.0016 & 0.0024 & 0.0008 \\
\cline{2-5}
& Mean & 0.0016 & 0.0024 & 0.0008 \\
\bottomrule
\end{tabular}
\renewcommand{\arraystretch}{1}
\caption{
[\textbf{Calibration} on \textit{self-consistency} scores] Using outputs from \textbf{Llama 2 7B Chat} on \textsc{Bio-NQ}, we calculate the ASCE for each group using \textbf{PS} and \textbf{GCULR} as well as the difference in ASCE ($\Delta$) between the two methods. We then present the top and bottom 5 groups according ($\Delta$) where \textbf{top} corresponds to groups for which the multivalid method achieves the biggest improvement (most negative change $\Delta$). In addition, we calculate the mean values for the top and bottom 5. We observe that GCULR greatly improves over PS among the top 5 groups, and even in the cases where GCULR worsens ASCE compared to PS, we find that the errors are already extremely small for both PS and GCULR.
}
\label{tab:asce_reg_best_worst_llama2}
\end{table*}


\begin{table*}[btp!]
\centering
\renewcommand{\arraystretch}{1.1}
\begin{tabular}{l l c c c}
\toprule
& & \textbf{PS} & \textbf{GCULR} & $\Delta$ \\
\toprule
\multirow{6}{*}{Top 5 $\Delta$}
& \textbf{\# Wiki prop.} = \textit{Low} \textcolor{ForestGreen}{\&} \textbf{nationality} = \textit{EU/ME} & 0.0573 & 0.0049 & -0.0524 \\
& \textbf{\# Wiki prop.} = \textit{Low} \textcolor{ForestGreen}{\&} \textbf{has IMDb ID} = \textit{True} & 0.0488 & 0.0035 & -0.0453 \\
& \textbf{\# Wiki prop.} = \textit{Low} \textcolor{ForestGreen}{\&} \textbf{sport} = \textit{False} & 0.0291 & 0.0011 & -0.0280 \\
& \textbf{\# Wiki prop.} = \textit{Low} \textcolor{ForestGreen}{\&} \textbf{nationality} = \textit{NA} & 0.0253 & 0.0018 & -0.0235 \\
& \textbf{\# Wiki prop.} = \textit{Low} \textcolor{ForestGreen}{\&} \textbf{sex or gender} = \textit{female} & 0.0226 & 0.0026 & -0.0200 \\
\cline{2-5}
& Mean & 0.0366 & 0.0028 & -0.0338 \\
\midrule
\multirow{6}{*}{Bottom 5 $\Delta$}
& \textbf{\# Wiki prop.} = \textit{Medium} \textcolor{ForestGreen}{\&} \textbf{nationality} = \textit{APAC} & 0.0018 & 0.0019 & 0.0001 \\
& \textbf{nationality} = \textit{APAC} \textcolor{ForestGreen}{\&} \textbf{sex or gender} = \textit{female} & 0.0020 & 0.0027 & 0.0008 \\
& \textbf{has IMDb ID} = \textit{False} \textcolor{ForestGreen}{\&} \textbf{sex or gender} = \textit{female} & 0.0011 & 0.0022 & 0.0011 \\
& \textbf{has IMDb ID} = \textit{True} \textcolor{ForestGreen}{\&} \textbf{nationality} = \textit{APAC} & 0.0010 & 0.0022 & 0.0012 \\
& \textbf{\# Wiki prop.} = \textit{Low} \textcolor{ForestGreen}{\&} \textbf{sport} = \textit{True} & 0.0021 & 0.0041 & 0.0020 \\
\cline{2-5}
& Mean & 0.0016 & 0.0026 & 0.0011 \\
\bottomrule
\end{tabular}
\renewcommand{\arraystretch}{1}
\caption{
[\textbf{Calibration} on \textit{self-consistency} scores] Using outputs from \textbf{Mistral 7B Instruct} on \textsc{Bio-NQ}, we calculate the ASCE for each group using \textbf{PS} and \textbf{GCULR} as well as the difference in ASCE ($\Delta$) between the two methods. We then present the top and bottom 5 groups according ($\Delta$) where \textbf{top} corresponds to groups for which the multivalid method achieves the biggest improvement (most negative change $\Delta$). In addition, we calculate the mean values for the top and bottom 5. We observe that GCULR greatly improves over PS among the top 5 groups, and even in the cases where GCULR worsens ASCE compared to PS, we find that the errors are already extremely small for both PS and GCULR.
}
\label{tab:asce_reg_best_worst_mistral}
\end{table*}

In Section~\ref{sec:results}, we demonstrate that multicalibration methods (IGHB, GCULR) significantly outperform standard calibration methods (HB, PS) with respect to calibration error both marginally and and within groups (max and mean gASCE). In this section, we further examine group calibration error, specifically looking at which groups do multicalibration methods improve over marginal methods most. 

First, in Tables \ref{tab:asce_hb_best_worst_llama2} and \ref{tab:asce_hb_best_worst_mistral}, we compare HB to IGHB for outputs from \textsc{Llama 2 7B Chat} and \textsc{Mistral 8B Instruct}, calculating the difference $\Delta$ in ASCE between the two methods for each group. Interestingly, we find that IGHB improves over HB for every group. We note that this finding is expected when one has access to the true data distribution. In our case, we implement IGHB using the calibration set (since we do not have access to the true data distribution), suggesting that the distributions for our calibration and test sets are still close enough such that IGHB is able to achieve such a strong result. Therefore, we present in Tables \ref{tab:asce_hb_best_worst_llama2} and \ref{tab:asce_hb_best_worst_mistral} the top 5 groups in terms of improvement $\Delta$ of IGHB compared to HB. For reference, we also present results for the group with the smallest improvement (to show the minimum improvement of the method).

Next, in Tables \ref{tab:asce_reg_best_worst_llama2} and \ref{tab:asce_reg_best_worst_mistral}, we compare PS to GCULR for outputs from \textsc{Llama 2 7B Chat} and \textsc{Mistral 8B Instruct}. Upon initial inspection, we find that unlike for HB and IGHB, GCULR does not improve ASCE for every single group when compared to its standard variant, PS. Thus, in Tables \ref{tab:asce_hb_best_worst_llama2} and \ref{tab:asce_hb_best_worst_mistral}, we instead show the top and bottom 5 groups in terms of improvement $\Delta$ of GCULR over PS. As shown in these results, like IGHB, GCULR is able to improve ASCE by a large margin (top 5 $\Delta$). Moreover, we find that among groups (bottom 5 $\Delta$) where GCULR is not able to improve ASCE, the calibration errors of PS are already very small ($\le 0.0028$ for Llama 2 and $\le 0.0021$ for Mistral). While GCULR does worsen ASCE for these 5 groups, the mean difference $\Delta$ is only $0.0008$ and $0.0011$ for \textsc{Llama 2 7B Chat} and \textsc{Mistral 8B Instruct}, thereby achieving still small calibration errors. In comparison, when GCULR does correct ASCE for subgroups, it does so by large margin, with mean reduction in error of $0.0306$ and $0.0338$ respectively. Consequently, we still see large improvements for overall mean and max gASCE when comparing GCULR to PS (as shown in Table~\ref{tab:multicalibration_asce_nq} of the main body).

Finally, we note that in all tables Tables \ref{tab:asce_hb_best_worst_llama2}, \ref{tab:asce_hb_best_worst_mistral}, \ref{tab:asce_reg_best_worst_llama2}, and \ref{tab:asce_reg_best_worst_mistral}, we observe the same set of groups in top and bottom 5, sorted by difference in ASCE $\Delta$. For example, regardless of model (Llama vs Mistral) or algorithm type (iterative patching vs regression-based), the top 5 groups (and their order) are exactly the same. Similarly, we find that for both models, the group with the smallest improvement is \textbf{\# Wiki prop.} = \textit{Medium} \textcolor{ForestGreen}{\&} \textbf{nationality} = \textit{APAC}. These observations suggest that our findings are not unique to either the model choice or calibration algorithm type.

Looking specifically at which groups does multicalibration correct the most (top 5 $\Delta$), we see that our models are most miscalibrated w.r.t. groups where the \# Wikidata properties is low, suggesting that standard calibration methods (HB, PS) are miscalibrated when it comes to quantifying uncertainty for individuals whose information is not prevalent on the Internet (and therefore most likely do not appear as often in training data used to train LLMs today). Fortunately, however, incorporating group information (as is done in IGHB and GCULR) helps alleviate this issue (i.e., in Tables \ref{tab:asce_hb_best_worst_llama2} and \ref{tab:asce_hb_best_worst_mistral}, the mean ASCE of GCULR for the top and bottom groups is fairly close).

\paragraph{Conformal Prediction.}

\begin{figure}[ bp!]
    \centering
    \begin{subfigure}[t]{0.75\textwidth}
        \centering
        \includegraphics[width=\textwidth]{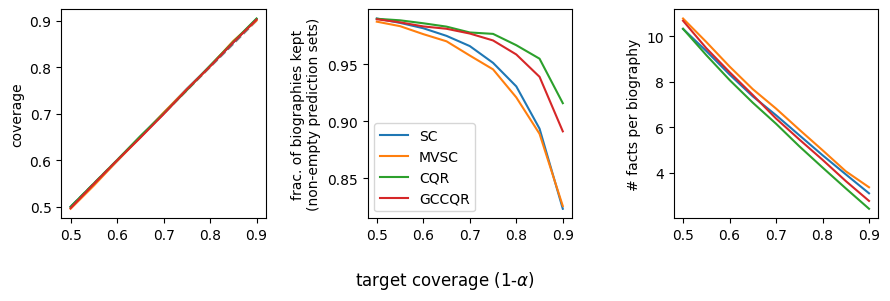}
        \caption{LLama 2 7B Chat}
    \end{subfigure}
    \begin{subfigure}[t]{0.75\textwidth}
        \centering
        \includegraphics[width=\textwidth]{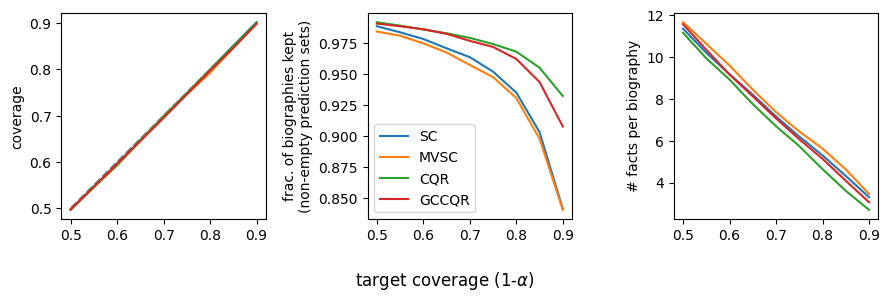}
        \caption{Mistral 7B Instruct}
    \end{subfigure}
    \caption{We report additional metrics for conformal predictions techniques when evaluated on biographies generated for \textsc{Bio-NQ}. Here, we use \textit{self-consistency} as our base uncertainty score function. On the \textbf{top} row, we present these metrics for outputs from LLama 2 7B Chat, and on the \textbf{bottom}, Mistral 7B Instruct. On the \textbf{left} panel, we plot the empirical coverage achieved against the target coverage. On the \textbf{middle} panel, we plot the fraction of biographies retained for each method against the target coverage level. Finally, on the \textbf{right} panel, we plot the number of facts per biography retained, again against the target coverage level.}
    \label{fig:conformal_nq_additional}
\end{figure}

In Figure \ref{fig:conformal_nq_additional}, we provide additional information about the prediction sets outputted by our various conformal methods on \textsc{Bio-NQ}. On the left panel, we plot the empirical coverage achieved against the target coverage. Figure \ref{fig:conformal_nq_additional} demonstrates that all methods achieve the target (marginal) coverage. On the \textbf{middle} panel, we plot the fraction of biographies retained (i.e., non-abstentions) for each method against the target coverage level, while on the \textbf{right} panel, we plot the number of facts per biography retained. Generally speaking, all methods retain about the same number of facts per biography. We also observe that to achieve the same target coverage, with SC and MVSC generally retaining fewer biographies (i.e., more abstentions) when compared to CQR and GCCQR. However, when comparing each conformal method (SC and CQR) to their multivalid counterparts (MVSC and GCCQR), we again observe that there are very little differences between them.

To help illustrate how different conformal methods (e.g., standard conformal vs. the multivalid counterpart) affect the final output text (i.e., subsets of retained claims), we provide examples\footnote{Note that these examples are meant to be illustrative---measuring actual effectiveness of conformal prediction methods must done at the group or dataset level (e.g., Figure \ref{fig:conformal_nq}).} outputted by models on \textsc{Bio-NQ}. In Tables \ref{tab:examples_llama2_retain_more} and \ref{tab:examples_mistral_retain_more}, we demonstrate how multivalid conformal methods can produce sets with additional claims retained. Moreover, in some cases, standard conformal methods (SC, CQR) may produce empty sets (abstain) while their multivalid counterparts do not (Tables \ref{tab:examples_llama2_nonempty} and \ref{tab:examples_mistral_nonempty}).

\begin{table*}[btp!]
\centering
\renewcommand{\arraystretch}{1.1}
\begin{tabular}{l l c c c}
\toprule
& & \textbf{SC} & \textbf{MVSC} & $\Delta$ \\
\toprule
\multirow{6}{*}{Top 5 $\Delta$}
& \textbf{\# Wiki prop.} = \textit{Very High} \textcolor{ForestGreen}{\&} \textbf{has IMDb ID}     = \textit{True}    & 0.0318 & 0.0137 & -0.0182 \\
& \textbf{\# Wiki prop.} = \textit{Low}       \textcolor{ForestGreen}{\&} \textbf{nationality}     = \textit{NA}      & 0.0325 & 0.0153 & -0.0171 \\
& \textbf{\# Wiki prop.} = \textit{High}      \textcolor{ForestGreen}{\&} \textbf{sex or gender}   = \textit{Female}  & 0.0312 & 0.0183 & -0.0129 \\
& \textbf{has IMDb ID}   = \textit{True}      \textcolor{ForestGreen}{\&} \textbf{plays pro sport} = \textit{True}    & 0.0427 & 0.0298 & -0.0129 \\
& \textbf{\# Wiki prop.} = \textit{Very High}                                                                         & 0.0256 & 0.0131 & -0.0125 \\
\cline{2-5}
& Mean                                                                                                                & 0.0328 & 0.0180 & -0.0147 \\
\midrule
\multirow{6}{*}{Bottom 5 $\Delta$}
& \textbf{has IMDb ID}   = \textit{False}     \textcolor{ForestGreen}{\&} \textbf{nationality}     = \textit{APAC}   & 0.0272 & 0.0401 & 0.0129 \\
& \textbf{\# Wiki prop.} = \textit{High}      \textcolor{ForestGreen}{\&} \textbf{has IMDb ID}     = \textit{False}  & 0.0180 & 0.0292 & 0.0112 \\
& \textbf{\# Wiki prop.} = \textit{Low}       \textcolor{ForestGreen}{\&} \textbf{has IMDb ID}     = \textit{True}   & 0.0183 & 0.0216 & 0.0033 \\
& \textbf{\# Wiki prop.} = \textit{Very High} \textcolor{ForestGreen}{\&} \textbf{has IMDb ID}     = \textit{False}  & 0.0259 & 0.0283 & 0.0024 \\
& \textbf{nationality}   = \textit{APAC}      \textcolor{ForestGreen}{\&} \textbf{plays pro sport} = \textit{False}  & 0.0231 & 0.0244 & 0.0013 \\
\cline{2-5}
& Mean                                                                                                               & 0.0225 & 0.0287 & 0.0062 \\
\bottomrule
\end{tabular}
\renewcommand{\arraystretch}{1}
\caption{[\textbf{Conformal} on \textit{self-consistency} scores] Using outputs from \textbf{Llama 2 7B Chat} on \textsc{Bio-NQ}, we calculate the coverage error (for a target coverage of $90\%$) for each group using \textbf{SC} and \textbf{MVSC} as well as the difference in coverage error ($\Delta$) between the two methods. We then present the top and bottom 5 groups according ($\Delta$) where \textbf{top} corresponds to groups for which the multivalid method achieves the biggest improvement (most negative change $\Delta$). In addition, we calculate the mean values for the top and bottom 5.}
\label{tab:conformal_sc_best_worst_llama2}
\end{table*}


\begin{table*}[btp!]
\centering
\renewcommand{\arraystretch}{1.1}
\begin{tabular}{l l c c c}
\toprule
& & \textbf{SC} & \textbf{MVSC} & $\Delta$ \\
\toprule
\multirow{6}{*}{Top 5 $\Delta$}
& \textbf{\# Wiki prop.} = \textit{Very High} \textcolor{ForestGreen}{\&} \textbf{nationality}     = \textit{EU/ME}  & 0.0423 & 0.0235 & -0.0188 \\
& \textbf{\# Wiki prop.} = \textit{Low}       \textcolor{ForestGreen}{\&} \textbf{has IMDb ID}     = \textit{True}   & 0.0282 & 0.0112 & -0.0170 \\
& \textbf{\# Wiki prop.} = \textit{Very High} \textcolor{ForestGreen}{\&} \textbf{sex or gender}   = \textit{Male}   & 0.0335 & 0.0190 & -0.0145 \\
& \textbf{has IMDb ID}   = \textit{Medium}    \textcolor{ForestGreen}{\&} \textbf{plays pro sport} = \textit{True}   & 0.0404 & 0.0269 & -0.0135 \\
& \textbf{\# Wiki prop.} = \textit{Very High} \textcolor{ForestGreen}{\&} \textbf{has IMDb ID}     = \textit{False}  & 0.0371 & 0.0240 & -0.0131 \\
\cline{2-5}
& Mean                                                                                                               & 0.0363 & 0.0209 & -0.0154 \\
\midrule
\multirow{6}{*}{Bottom 5 $\Delta$}
& \textbf{nationality}     = \textit{NA}     \textcolor{ForestGreen}{\&} \textbf{plays pro sport} = \textit{True}    & 0.0173 & 0.0282 & 0.0109 \\
& \textbf{has IMDb ID}     = \textit{False}  \textcolor{ForestGreen}{\&} \textbf{plays pro sport} = \textit{True}    & 0.0221 & 0.0279 & 0.0058 \\
& \textbf{plays pro sport} = \textit{False}  \textcolor{ForestGreen}{\&} \textbf{sex or gender}   = \textit{Female}  & 0.0138 & 0.0192 & 0.0055 \\
& \textbf{sex or gender}   = \textit{Female}                                                                         & 0.0143 & 0.0193 & 0.0050 \\
& \textbf{has IMDb ID}     = \textit{True}   \textcolor{ForestGreen}{\&} \textbf{sex or gender}   = \textit{Female}  & 0.0145 & 0.0190 & 0.0044 \\
\cline{2-5}
& Mean                                                                                                               & 0.0164 & 0.0227 & 0.0063 \\
\bottomrule
\end{tabular}
\renewcommand{\arraystretch}{1}
\caption{[\textbf{Conformal} on \textit{self-consistency} scores] Using outputs from \textbf{Mistral 7B Instruct} on \textsc{Bio-NQ}, we calculate the coverage error (for a target coverage of $90\%$) for each group using \textbf{SC} and \textbf{MVSC} as well as the difference in coverage error ($\Delta$) between the two methods. We then present the top and bottom 5 groups according ($\Delta$) where \textbf{top} corresponds to groups for which the multivalid method achieves the biggest improvement (most negative change $\Delta$). In addition, we calculate the mean values for the top and bottom 5.}
\label{tab:conformal_sc_best_worst_mistral}
\end{table*}


\begin{table*}[btp!]
\centering
\renewcommand{\arraystretch}{1.1}
\begin{tabular}{l l c c c}
\toprule
& & \textbf{CQR} & \textbf{GCCQR} & $\Delta$ \\
\toprule
\multirow{6}{*}{Top 5 $\Delta$}
& \textbf{\# Wiki prop.} = \textit{Low} \textcolor{ForestGreen}{\&} \textbf{sport}         = \textit{False} & 0.0652 & 0.0190 & -0.0463 \\
& \textbf{\# Wiki prop.} = \textit{Low} \textcolor{ForestGreen}{\&} \textbf{IMDb ID}       = \textit{True}  & 0.0564 & 0.0153 & -0.0411 \\
& \textbf{\# Wiki prop.} = \textit{Low}                                                                     & 0.0556 & 0.0167 & -0.0389 \\
& \textbf{\# Wiki prop.} = \textit{Low} \textcolor{ForestGreen}{\&} \textbf{IMDb ID}       = \textit{False} & 0.0565 & 0.0226 & -0.0339 \\
& \textbf{\# Wiki prop.} = \textit{Low} \textcolor{ForestGreen}{\&} \textbf{sex or gender} = \textit{Male}  & 0.0556 & 0.0222 & -0.0334 \\
\cline{2-5}
& Mean                                                                                                      & 0.0579 & 0.0192 & -0.0387 \\
\midrule
\multirow{6}{*}{Bottom 5 $\Delta$}
& \textbf{nationality}   = \textit{APAC}      \textcolor{ForestGreen}{\&} \textbf{sport}         = \textit{False}  & 0.0150 & 0.0294 & 0.0143 \\
& \textbf{\# Wiki prop.} = \textit{Very High} \textcolor{ForestGreen}{\&} \textbf{IMDb ID}       = \textit{False}  & 0.0193 & 0.0335 & 0.0142 \\
& \textbf{nationality}   = \textit{APAC}      \textcolor{ForestGreen}{\&} \textbf{sex or gender} = \textit{Male}   & 0.0151 & 0.0227 & 0.0076 \\
& \textbf{\# Wiki prop.} = \textit{Medium}    \textcolor{ForestGreen}{\&} \textbf{nationality}   = \textit{EU/ME}  & 0.0196 & 0.0268 & 0.0072 \\
& \textbf{\# Wiki prop.} = \textit{Medium}    \textcolor{ForestGreen}{\&} \textbf{sex or gender} = \textit{Female} & 0.0183 & 0.0241 & 0.0058 \\
\cline{2-5}
& Mean                                                                                                             & 0.0175 & 0.0273 & 0.0098 \\
\bottomrule
\end{tabular}
\renewcommand{\arraystretch}{1}
\caption{[\textbf{Conformal} on \textit{self-consistency} scores] Using outputs from \textbf{Llama 2 7B Chat} on \textsc{Bio-NQ}, we calculate the coverage error (for a target coverage of $90\%$) for each group using \textbf{CQR} and \textbf{GCCQR} as well as the difference in coverage error ($\Delta$) between the two methods. We then present the top and bottom 5 groups according ($\Delta$) where \textbf{top} corresponds to groups for which the multivalid method achieves the biggest improvement (most negative change $\Delta$). In addition, we calculate the mean values for the top and bottom 5.}
\label{tab:conformal_qreg_best_worst_llama2}
\end{table*}


\begin{table*}[btp!]
\centering
\renewcommand{\arraystretch}{1.1}
\begin{tabular}{l l c c c}
\toprule
& & \textbf{CQR} & \textbf{GCCQR} & $\Delta$ \\
\toprule
\multirow{6}{*}{Top 5 $\Delta$}
& \textbf{\# Wiki prop.} = \textit{Low} \textcolor{ForestGreen}{\&} \textbf{nationality}   = \textit{NA}    & 0.0795 & 0.0135 & -0.0661 \\
& \textbf{\# Wiki prop.} = \textit{Low} \textcolor{ForestGreen}{\&} \textbf{IMDb ID}       = \textit{True}  & 0.0758 & 0.0173 & -0.0585 \\
& \textbf{\# Wiki prop.} = \textit{Low} \textcolor{ForestGreen}{\&} \textbf{sport}         = \textit{False} & 0.0746 & 0.0175 & -0.0571 \\
& \textbf{\# Wiki prop.} = \textit{Low}                                                                     & 0.0662 & 0.0121 & -0.0541 \\
& \textbf{\# Wiki prop.} = \textit{Low} \textcolor{ForestGreen}{\&} \textbf{sex or gender} = \textit{Male}  & 0.0698 & 0.0244 & -0.0454 \\
\cline{2-5}
& Mean                                                                                                      & 0.0732 & 0.0170 & -0.0562 \\
\midrule
\multirow{6}{*}{Bottom 5 $\Delta$}
& \textbf{IMDb ID}       = \textit{False}  \textcolor{ForestGreen}{\&} \textbf{nationality}   = \textit{EU/ME}  & 0.0193 & 0.0331 & 0.0138 \\
& \textbf{nationality}   = \textit{NA}     \textcolor{ForestGreen}{\&} \textbf{sport}         = \textit{True}   & 0.0155 & 0.0283 & 0.0128 \\
& \textbf{IMDb ID}       = \textit{False}  \textcolor{ForestGreen}{\&} \textbf{sport}         = \textit{False}  & 0.0114 & 0.0194 & 0.0080 \\
& \textbf{nationality}   = \textit{EU/ME}  \textcolor{ForestGreen}{\&} \textbf{sex or gender} = \textit{Female} & 0.0275 & 0.0352 & 0.0077 \\
& \textbf{IMDb ID}       = \textit{False}  \textcolor{ForestGreen}{\&} \textbf{sport}         = \textit{True}   & 0.0267 & 0.0325 & 0.0057 \\
\cline{2-5}
& Mean                                                                                                          & 0.0201 & 0.0297 & 0.0096 \\
\bottomrule
\end{tabular}
\renewcommand{\arraystretch}{1}
\caption{[\textbf{Conformal} on \textit{self-consistency} scores] Using outputs from \textbf{Mistral 7B Instruct} on \textsc{Bio-NQ}, we calculate the coverage error (for a target coverage of $90\%$) for each group using \textbf{CQR} and \textbf{GCCQR} as well as the difference in coverage error ($\Delta$) between the two methods. We then present the top and bottom 5 groups according ($\Delta$) where \textbf{top} corresponds to groups for which the multivalid method achieves the biggest improvement (most negative change $\Delta$). In addition, we calculate the mean values for the top and bottom 5.}
\label{tab:conformal_qreg_best_worst_mistral}
\end{table*}

Finally, like in the section above, we again examine what groups do multivalid conformal methods improve over standard methods on the most, where in this case, we instead calculate the difference $\Delta$ in coverage error (at target coverage of $90\%$) between each pairing of conformal and multivalid conformal methods. In particular, we display the top and bottom 5 groups in terms of difference $\Delta$ in Tables \ref{tab:conformal_sc_best_worst_llama2}, \ref{tab:conformal_sc_best_worst_mistral}, \ref{tab:conformal_qreg_best_worst_llama2}, and \ref{tab:conformal_qreg_best_worst_mistral}

Our findings show that for the topic of factuality in long-form text generation, multivalid conformal prediction is a more challenging problem when compared to calibration. As shown in Figure \ref{fig:conformal_nq}, multivalid methods (GCCQR and MVSC) consistently (at all target coverages) outperform standard conformal methods (SC and CQR) w.r.t. group coverage error. Tables \ref{tab:conformal_sc_best_worst_llama2}, \ref{tab:conformal_sc_best_worst_mistral}, \ref{tab:conformal_qreg_best_worst_llama2}, and \ref{tab:conformal_qreg_best_worst_mistral} corroborate this finding, showing that the mean coverage difference $\Delta$ for the top groups is larger (at a minimum, $2.41$x more for MVSC and $3.95$x more for GCCQR), demonstrating that multivalid methods tend to improve coverage error on groups more than it worsens it (for other groups), thereby achieving a better mean group coverage error overall. However, the improvements are not as stark as those found in Tables \ref{tab:asce_hb_best_worst_llama2}, \ref{tab:asce_hb_best_worst_mistral}, \ref{tab:asce_reg_best_worst_llama2}, and \ref{tab:asce_reg_best_worst_mistral} for calibration error, suggesting that multivalid conformal prediction may be a harder problem overall.

When looking at which groups do multivalid conformal methods improve the most on, we find no consistent patterns. However, we do observe that all groups for which MVSC or GCCQR improve the most on are related to the number of Wikidata properties. Interesting, we do observe that CQR does quite poorly on groups containing people with a low number of Wikidata properties, mirroring our findings for calibration above. Like in multicalibration, GCCQR is able to significanly improve coverage error for these groups. Lastly, we note that CQR seems to achieve worse group coverage than that of SC. For example, on outputs from \textsc{Mistral 7B Instruct}, the mean coverage error among the top 5 groups is $0.0732$ for CQR compared to $0.0363$ for SC. However, we find that for both models (Tables \ref{tab:conformal_qreg_best_worst_llama2}, and \ref{tab:conformal_qreg_best_worst_mistral}), GCCQR is able to still reduce coverage errors to levels similar to that of MVSC (Tables \ref{tab:conformal_sc_best_worst_llama2}, and \ref{tab:conformal_sc_best_worst_mistral}).

\begin{table*}[btp!]
\centering
\renewcommand{\arraystretch}{1.2}
\begin{tabular}{p{4in}|cc|cc}
\toprule
 Claims & SC & MVSC & CQR & GCCQR \\
\toprule
Henry Cavill was born in Jersey, Channel Islands. & X & X & X & X \\
Henry Cavill has reprised the role of Superman in "Batman v Superman: Dawn of Justice" (2016). &  & X & X & X \\
Henry Cavill gained international recognition for his portrayal of Superman in the DC Extended Universe. &  & X &  & X \\
Henry Cavill has reprised the role of Superman in "Justice League" (2017). &  & X &  & X \\
Henry Cavill is British. &  &  &  & X \\
Henry Cavill is also known for his philanthropic work. &  &  &  & X \\
Henry Cavill is an actor. &  &  &  & X \\
Henry Cavill was born on May 5th, 1983. &  &  &  & X \\
Henry Cavill's performance in the role of Superman has been widely praised. &  &  &  & X \\
\midrule
Amy Winehouse left a lasting impact on the music industry. & X & X & X & X \\
Amy Winehouse released her follow-up album, "Back to Black," in 2006. & X & X & X & X \\
Amy Winehouse's debut album "Frank" was released in 2003. & X & X & X & X \\
Amy Winehouse was a unique artist. & X & X &  & X \\
Amy Winehouse's lyrics were poignant. & X & X &  & X \\
The hit single "Rehab" contributed to the album's success. & X & X &  & X \\
Winehouse began singing and writing songs at a young age. & X & X &  & X \\
"Frank" received critical acclaim. &  & X &  &  \\
Amy Winehouse was a British singer and songwriter. &  & X &  &  \\
Amy Winehouse was a talented singer-songwriter. &  & X &  &  \\
Fans mourned the loss of Amy Winehouse. &  & X &  &  \\
Winehouse grew up in a family of Jewish descent. &  & X &  &  \\
\bottomrule
\end{tabular}
\renewcommand{\arraystretch}{1}
\caption{Using outputs from \textbf{Llama 2 7B Chat} on \textsc{Bio-NQ}, we present examples in which all conformal methods using \textit{self-consistency} scores (at $90\%$ target coverage) produce a subset of claims that are entirely correct. In these examples, multivalid methods (MVSC, GCCQR) retain more claims.}
\label{tab:examples_llama2_retain_more}
\end{table*}


\begin{table*}[btp!]
\centering
\renewcommand{\arraystretch}{1.2}
\begin{tabular}{p{4in}|cc|cc}
\toprule
 Claims & SC & MVSC & CQR & GCCQR \\
\toprule
H. G. Wells is considered a pioneer of science fiction. & X & X & X & X \\
H.G. Wells is best remembered for H. G. Wells's works in the science fiction genre. & X & X & X & X \\
H.G. Wells is most famous for H. G. Wells's science fiction. & X & X & X & X \\
H.G. Wells was a pioneer of the science fiction genre. & X & X & X & X \\
H.G. Wells was a prolific writer. & X & X & X & X \\
H.G. Wells was born in Bromley, England. & X & X & X & X \\
H.G. Wells was known for H. G. Wells's science fiction works. & X & X & X & X \\
H.G. Wells' most famous works include "The Time Machine," "The War of the Worlds," and "The Invisible Man." & X & X & X & X \\
H.G. Wells was a renowned writer. & X & X &  & X \\
H.G. Wells was an English writer. &  & X &  & X \\
H.G. Wells died in 1946. &  & X &  &  \\
H.G. Wells was a prolific writer, publishing over 50 books. &  & X &  &  \\
H.G. Wells was born on September 21, 1866, in Bromley, Kent, England. &  & X &  &  \\
H.G. Wells wrote works in various other genres, including fiction. &  & X &  &  \\
H.G. Wells wrote works in various other genres, including social commentary. &  & X &  &  \\
H.G. Wells' works often explored the social and political implications of scientific and technological advancements. &  & X &  &  \\
Wells' works, such as "The Time Machine," "The War of the Worlds," and "The Invisible Man," have had a significant impact on the development of the science fiction literary genre. &  & X &  &  \\
\midrule
Heisenberg made significant contributions to quantum mechanics. & X & X & X & X \\
Werner Heisenberg passed away on February 1, 1976. & X & X & X & X \\
Werner Heisenberg studied under Arnold Sommerfeld at the University of Munich. & X & X & X & X \\
Werner Heisenberg was a key figure in the development of quantum mechanics. & X & X & X & X \\
Werner Heisenberg was born in Wurzberg, Germany in 1901. & X & X & X & X \\
Werner Heisenberg's work had a profound impact on the field of physics. & X & X & X & X \\
Heisenberg played a pioneering role in quantum theory. &  & X &  & X \\
Werner Heisenberg attended the University of Munich. &  & X &  & X \\
Werner Heisenberg is best known for his uncertainty principle. &  & X &  & X \\
Werner Heisenberg's contributions paved the way for the development of quantum mechanics. &  & X &  & X \\
Werner Heisenberg's work revolutionized the field of physics. &  & X &  & X \\
\bottomrule
\end{tabular}
\renewcommand{\arraystretch}{1}
\caption{Using outputs from \textbf{Mistral 7B Instruct} on \textsc{Bio-NQ}, we present examples in which all conformal methods using \textit{self-consistency} (at $90\%$ target coverage) produce a subset of claims that are entirely correct. In these examples, multivalid methods (MVSC, GCCQR) retain more claims.}
\label{tab:examples_mistral_retain_more}
\end{table*}
\begin{table*}[btp!]
\centering
\renewcommand{\arraystretch}{1.2}
\begin{tabular}{p{4in}|cc|cc}
\toprule
 & SC & MVSC & CQR & GCCQR \\
\toprule
Anaximander believed that the universe is infinite. &  & X &  & X \\
Anaximander came from a noble family. &  & X &  & X \\
Anaximander was born in Miletus, a city in the ancient Greek world. &  & X &  & X \\
Anaximander's work has survived to the present day. &  & X &  & X \\
Despite his contributions to philosophy, Anaximander's life remains somewhat shrouded in mystery. &  & X &  & X \\
\midrule
Merritt Butrick is best known for his roles in the Star Trek franchise. &  & X &  & X \\
Merritt Butrick was an American actor. &  & X &  & X \\
Merritt Butrick contributed to the Star Trek franchise. &  &  &  & X \\
\bottomrule
\end{tabular}
\caption{Using outputs from \textbf{Llama 2 7B Chat} on \textsc{Bio-NQ}, we present examples in which all conformal methods (at $90\%$ target coverage) produce a subset of claims that are entirely correct. In these examples, the multivalid methods (MVSC, GCCQR) output nonempty steps while the standard conformal methods (SC, CQR) do not.}
\label{tab:examples_llama2_nonempty}
\renewcommand{\arraystretch}{1}
\end{table*}


\begin{table*}[btp!]
\centering
\renewcommand{\arraystretch}{1.2}
\begin{tabular}{p{4in}|cc|cc}
\toprule
 & SC & MVSC & CQR & GCCQR \\
\toprule

Bessel van der Kolk has written extensively on the connection between the brain, mind, and body in the healing of trauma. &  & X & X & X \\
Bessel van der Kolk is a world-renowned Dutch-American psychiatrist. &  & X & X & X \\
Bessel van der Kolk's work has had a significant impact on the understanding and treatment of trauma. &  & X &  & X \\
\midrule
Richard Chamberlain continues to act. &  &  &  & X \\
Richard Chamberlain has received numerous awards and accolades throughout his career. &  &  &  & X \\
Richard Chamberlain was born on March 31, 1934, in Beverly Hills, California. &  &  &  & X \\
\bottomrule
\end{tabular}
\caption{Using outputs from \textbf{Mistral 7B Instruct} on \textsc{Bio-NQ}, we present examples in which all conformal methods using \textit{self-consistency} (at $90\%$ target coverage) produce a subset of claims that are entirely correct. In these examples, we have that either SC or CQR produce empty sets while their multivalid counterparts (MVSC and GCCQR respectively) do not.}
\label{tab:examples_mistral_nonempty}
\renewcommand{\arraystretch}{1}
\end{table*}


\clearpage

\section{Additional experimental details}\label{appx:additional_exp_details}


\subsection{Biography generation and factuality evaluation}\label{appx:generating}

While the ground truth score must be human-annotated, \citet{min2023factscore} show that \textsc{FActSCORE} can be approximated by an automated process that leverages an LLM (i.e., ChatGPT and LLaMa-7B) and natural language retrieval. Following \citet{min2023factscore}, we also use an LLM to automate the annotation process. For some input person, \textbf{\texttt{[ENTITY]}}, we prompt a large language model with the following:

\noindent \textbf{\texttt{[INST] Question: Tell me a bio of [ENTITY]. [/INST]}}

We then decompose each long-form generation into a set of atomic facts, which are then checked against some set of Wikipedia articles about the \textbf{\texttt{[ENTITY]}} to evaluate overall performance of language model in terms of factuality. \citet{min2023factscore} demonstrate that while the evaluation process should ideally be conducted by human annotators, using large language models (i.e., ChatGPT and LLama 1 7B) to both decompose long-form generations and check against Wikipedia articles serves as a very good proxy for human annotation.

Following this general framework for automated evaluation, we use Llama 2 7B Chat to decompose each generation \textbf{\texttt{[GEN\_BIO]}} with the following prompt:

\noindent \textbf{\texttt{[INST] <<SYS>> Break down the following input into a set of small, independent claims. You must not add additional information. Output the claims as a numbered list separated by a new line. The subject of each line should be [ENTITY]. <</SYS>> Input: [GEN\_BIO] [/INST] }}

For checking each atomic fact against Wikpedia, we directly use the code released by \citet{min2023factscore}, which first conducts passage retrieval via Generalizable T5-based Retrievers \citep{liu2023evaluating} to find relevant articles from a dump of Wikipedia (dated 2023-04-01) and then prompts an LLM (i.e., ChatGPT or Llama 1 7B) to predict whether each fact is supported by the retrieved passages. For our evaluation, we again use Llama 2 7B Chat. Finally, these predictions are ensembled with predictions using likelihood estimates derived from a nonparametric masked language model \citep{min2023nonparametric}.

We note that for prompting the LLMs described above, we use \href{http://huggingface.co}{Hugging Face}'s transformer's library and generate responses with temperature set to $1.0$.


\subsection{Base scoring functions}\label{appx:base_scoring_fns}
    
\paragraph{Self-consistency.} Instead of manually annotating which claims are contained in the additional generations, we automate the process. Specifically, we use a procedure similar to frequency scoring algorithm proposed by \citet{wang2024fine} in which \textbf{(1)} a set of $K$ most relevant claims from a reference generation is retrieved using a vanilla BM25 algorithm (to reduce computational costs). Then \textbf{(2)} an LLM is tasked to evaluate whether the target claim is supported by the set of $K$ reference claims. In our work, we replace LLM prompting in step \textbf{(2)} with \textsc{AlignScore-Large} \citep{zha2023alignscore}, which runs significantly faster and is reported by \citet{zha2023alignscore} to compare favorably to LLM-based alignment methods.

\paragraph{P(True).} We use the following prompt:

\noindent \textbf{\texttt{[INST] <<SYS>> Answer the question based on your knowledge of the topic, [TOPIC]. If you are unsure about the question, output False. <</SYS>> Question: Is the following statement True or False? [CLAIM] [/INST]}}

\paragraph{Verbalized confidence.} We use the following prompt:

\noindent \textbf{\texttt{[INST] <<SYS>> Given a [TOPIC]: [CLAIM] pair as input, use your knowledge about [TOPIC] to rate (on an integer scale between 1 and 5) how confident you are that the input [CLAIM] is true. <</SYS>> [TOPIC]: [CLAIM] [/INST]}}


\subsection{Miscellaneous details}

\paragraph{Datasets.} Table \ref{tab:dataset_stats} reports additional information about datasets \textsc{Bio-NQ} and \textsc{Bio-FActScore}, including the number of entities and claims per biography outputted by each model. 

\begin{table*}[hbtp!]
\centering
\begin{tabular}{c c | c | c | c }
\toprule
\textbf{dataset} & \textbf{model}         & \textbf{\# entities} & \textbf{total \# claims} & \textbf{avg. \# claims per bio.} \\ 
\midrule
\multirow{2}{*}{\textsc{Bio-NQ}}
& Llama 2 7B Chat     & 8,541 & 206,620 & 24.19 \\
& Mistral 7B Instruct & 8,541 & 297,714 & 34.86 \\ 
\midrule
\multirow{2}{*}{\textsc{Bio-FAcTScore}} 
& Llama 2 7B Chat     & 683 & 17,605 & 25.78 \\
& Mistral 7B Instruct & 683 & 25,283 & 37.02 \\ 
\bottomrule
\end{tabular}
\caption{Statistics describing our two datasets and how many claims are generated by each LLM.}
\label{tab:dataset_stats}
\end{table*}

\paragraph{Group Attributes.}
For our experiments, we use the following group attributes:
\begin{itemize}[itemsep=-2pt]
    \item \textbf{\# Wikidata properties}: For each entity, we count the number of Wikidata properties and discretize them into the following buckets: $[0, 25), [25, 50), [50-100), [100, \infty)$. This group serves as proxy for the amount of information available online for some given entity.
    
    \item \textbf{nationality}: Following \citet{min2023factscore}, who use nationality derived from Wikidata to sample their dataset of human entities, we take the property \textit{country of citizenship} (or \textit{place of birth} when not available) and categorize the corresponding value into the following categories defined by \citet{min2023factscore}: Asia/Pacific, Europe/Middle East, North America, Latin/South America/Africa.
    
    \item \textbf{sex or gender}: We take directly the value for the Wikidata property, \textit{sex or gender}.

    \item \textbf{plays professional sports}: We check whether the Wikidata entry has the property, \textit{sport}.

    \item \textbf{has IMDb entry}: We check whether the Wikidata entry has the property, \textit{IMDb ID}, to use as a proxy for whether a person has been involved in films or television series.
\end{itemize}

In total, we have $|\mathcal{G}| = 77$ subgroups. To prevent extremely uncommon groups that may exist in the Wikidata database from biasing our results, we exclude groups of size $<5\%$ of the total test set size. Note that while we create groups using $1$ and $2$-way combinations for evaluation, we train the quantile regression models in CQR and GCCQR using only single attribute groups as features in order to reduce computation.

\paragraph{Hyperparameters.} For our patching algorithms IGHB and MVSC, we set the max iterations $T=100$. For training (multi)calibration, our logistic regression models are trained using default hyperparameters given my \href{https://scikit-learn.org/stable/modules/generated/sklearn.linear_model.LogisticRegression.html}{Sci-kit learn}. For training CQR and GCCQR, we run 5-fold cross validation for each target coverage $1-\alpha$ to optimize the $\ell_1$-penalty term $C \in \{ 10^{-6}, 10^{-5}, 10^{-4}, 10^{-3}, 10^{-2}, 10^{-1} \}$

For \textsc{AlignScore}, we set $M=4$ and $K=5$. We found that \textsc{AlignScore} generally returns values close to $0$ or $1$, giving us self-consistency uncertainty scores around the $5$ values $\{ 0, \frac{1}{4}, \frac{1}{2}, \frac{3}{4}, 1 \}$. As a result, we evaluate all methods using $p=5$ level sets.

\paragraph{GPU requirements.} We use a NVIDIA A100 80GB GPU for all experiments. For obtaining results on all entities across \textsc{Bio-NQ} and \textsc{Bio-FActScore}, our experiments, per LLM require approximately the following:
\begin{itemize}[noitemsep]
    \item Generating biographies (+ 4 additional generations for getting frequency scores): 15 hours (x5)
    \item Splitting atomic facts (+ 4 additional generations for getting frequency scores): 30 hours (x5)
    \item Checking facts against Wikipedia: 75 hours (x1)
    \item Calculating frequency scores via AlignScore: 10 hours (x1)
\end{itemize}

\paragraph{Licenses.} Wikidata and Wikipedia are licensed under the Creative Commons CC0 License. Llama 2 7B is licensed under Meta's Llama 2 license. Mistral 7B Chat and Hugging Face's transformers library are licensed under Apache 2.0 license. We also make use of \href{https://github.com/shmsw25/FActScore}{code} released by \citet{min2023factscore} under the MIT license.


\clearpage

\section{Evaluating on entities used in Min et. al (2023a)}

In addition to evaluating on our dataset, \textsc{Bio-NQ}, we construct an additional dataset using the $683$ entities used by \citet{min2023factscore} for their empirical evaluation. We denote this dataset as \textsc{Bio-FActScore} and evaluate all methods using \textit{self-consistency} as the base scoring function.

\paragraph{Calibration.} In Table \ref{tab:multicalibration_asce_fs}, we observe similar results to that on \textsc{Bio-NQ}---namely, multicalibrated counterparts (IGHB and GCULR) perform better than their base counterpart (HB and PS). However, we note that for Mistral 7B Instruct, PS performs the best when looking at marginal ASCE. We hypothesize that the smaller gap in ASCE between PS and GCULR may be due to the smaller training size of \textsc{Bio-FActScore} (25,283 claims), which is roughly 10x smaller than that of \textsc{Bio-NQ} (297,714 claims). Lastly, with respect to Brier score, multicalibration still dominates across all metrics (Table \ref{tab:multicalibration_brier_fs}).

\begin{table*}[b!]
\centering
\begin{tabular}{c c | c | c c | c c }
\toprule
\textbf{Model}        & \textbf{Metric}   & \textbf{Uncalibrated} & \textbf{HB}     & \textbf{IGHB}     & \textbf{PS}   & \textbf{GCULR} \\ 
\midrule
\multirow{3}{*}{Llama 2 7B Chat} 
& marginal   & 0.26830     & 0.00951   & \textbf{0.00229} & 0.00164 & \textbf{0.00125}* \\ 
& group max  & 0.48594     & 0.07208   & \textbf{0.04088} & 0.05017 & \textbf{0.03519}* \\ 
& group mean & 0.29983     & 0.02848   & \textbf{0.01108} & 0.01659 & \textbf{0.00858}* \\ 
\midrule
\multirow{3}{*}{Mistral 7B Instruct} 
& marginal   & 0.25496     & 0.01032   & \textbf{0.00268} & \textbf{0.00093}* & 0.00146 \\ 
& group max. & 0.54701     & 0.08436   & \textbf{0.04585}* & 0.07043 & \textbf{0.04931} \\ 
& group mean & 0.29435     & 0.03226   & \textbf{0.01143} & 0.01848 & \textbf{0.00911}* \\ 
\bottomrule
\end{tabular}
\caption{We generate biographies for entities from \textsc{Bio-FActScore} and compare each calibration method (HB, PS) against its multicalibration counterpart (IGHB, GCULR) on \textbf{ASCE}, \textbf{max gASCE}, and \textbf{average gASCE} ($\downarrow$ better). We bold the better-performing method for each pairing. * denotes the best-performing method across all methods evaluated. All methods use \textit{self-consistency} as their base scoring function.}
\label{tab:multicalibration_asce_fs}
\end{table*}
\begin{table*}[b!]
\centering
\begin{tabular}{c c | c | c c | c c }
\toprule
\textbf{Model}        & \textbf{Metric}   & \textbf{Uncalibrated} & \textbf{HB}     & \textbf{IGHB}     & \textbf{PS}   & \textbf{GCULR} \\ 
\midrule
\multirow{3}{*}{Llama 2 7B Chat} 
& marginal  & 0.475     & 0.169   & \textbf{0.148} & 0.152 & \textbf{0.143}* \\ 
& group max & 0.535     & 0.323   & \textbf{0.247} & 0.285 & \textbf{0.235}* \\ 
& group mean & 0.479     & 0.169   & \textbf{0.148} & 0.152 & \textbf{0.142}* \\ 
\midrule
\multirow{3}{*}{Mistral 7B Instruct} 
& marginal  & 0.471     & 0.186   & \textbf{0.159} & 0.164 & \textbf{0.152}* \\ 
& group max & 0.554     & 0.333   & \textbf{0.250} & 0.285 & \textbf{0.235}* \\ 
& group mean & 0.477     & 0.186   & \textbf{0.158} & 0.164 & \textbf{0.152}* \\ 
\bottomrule
\end{tabular}
\caption{We generate biographies for entities from \textsc{Bio-FActScore} and compare each calibration method (HB, PS) against its multicalibration counterpart (IGHB, GCULR) on \textbf{Brier score} ($\downarrow$ better) \textbf{marginally} across the entire dataset, as well as within each subgroup (in terms of \textbf{max} and \textbf{mean} over all groups). We bold the better-performing method for each pairing. * denotes the best-performing method across all methods evaluated. All methods use \textit{self-consistency} as their base scoring function.}
\label{tab:multicalibration_brier_fs}
\end{table*}

\begin{figure}[tbp!]
    \centering
    \begin{subfigure}{0.6\textwidth}
        \centering
        \includegraphics[width=\textwidth]{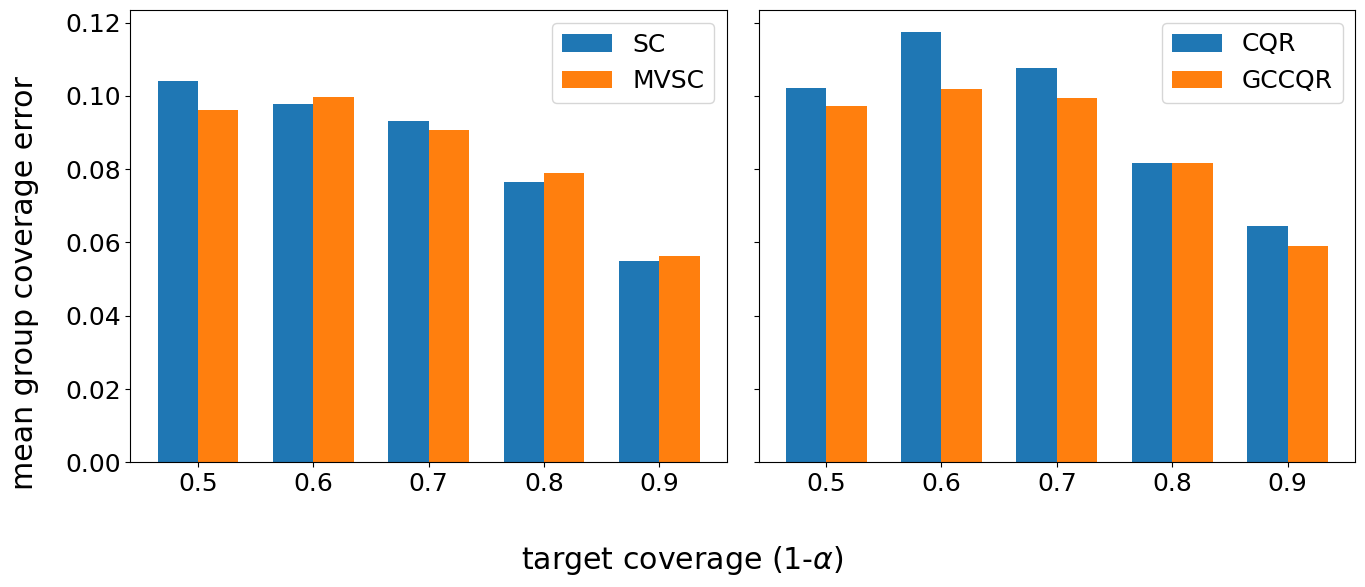}
        \caption{LLama 2 7B Chat}
        \label{fig:conformal_fs_llama}
    \end{subfigure}
    \begin{subfigure}{0.6\textwidth}
        \centering
        \includegraphics[width=\textwidth]{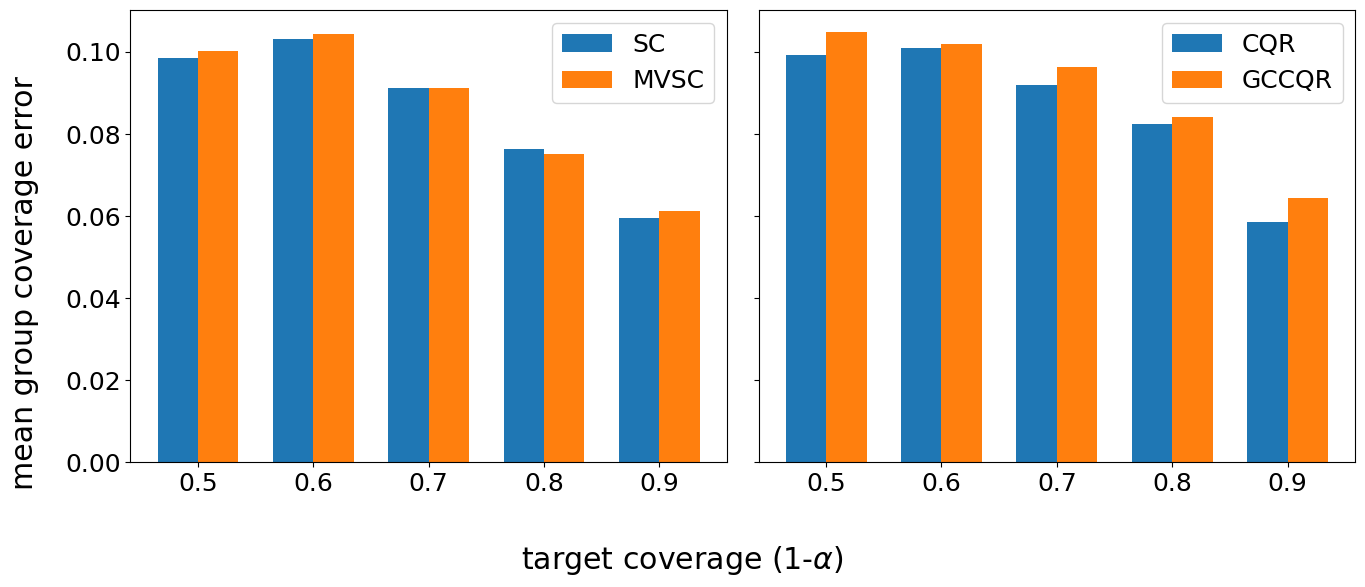}
        \caption{Mistral 7B Instruct}
        \label{fig:conformal_fs_mistral}
    \end{subfigure}
    \caption{For each target coverage, we run conformal methods (SC, CQR) and their multigroup counterparts (MVSC, GCCQR) on \textsc{Bio-FactScore}. We evaluate on generations by (a) Llama 2 7B Chat and (b) Mistral 7B Instruct. We calculate the average coverage error across all groups and plot them side by side for each pairing. All methods use \textit{self-consistency} as their base scoring function.}
    \label{fig:conformal_fs}
\end{figure}

\begin{figure}[tbp!]
    \centering
    \begin{subfigure}[t]{0.75\textwidth}
        \centering
        \includegraphics[width=\textwidth]{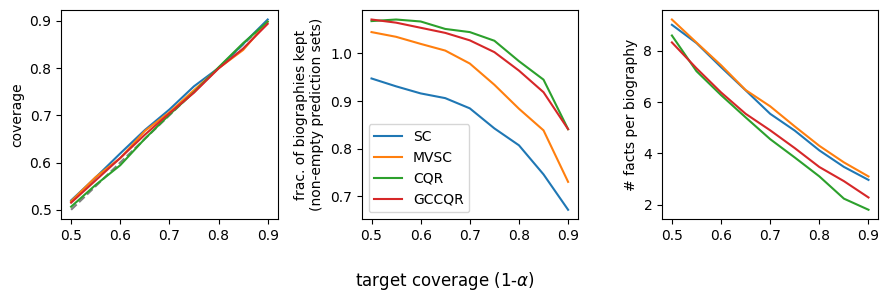}
        \caption{LLama 2 7B Chat}
    \end{subfigure}
    \begin{subfigure}[t]{0.75\textwidth}
        \centering
        \includegraphics[width=\textwidth]{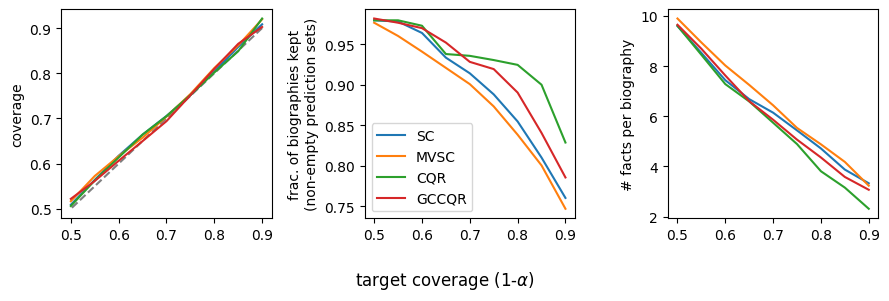}
        \caption{Mistral 7B Instruct}
    \end{subfigure}
    \caption{We report additional metrics for conformal predictions techniques using \textit{self-consistency} scores when evaluated on biographies generated for \textsc{Bio-FActScore}: On the \textbf{top} row, we present these metrics for outputs from LLama 2 7B Chat, and on the \textbf{bottom}, Mistral 7B Instruct. On the \textbf{left} panel, we plot the empirical coverage achieved against the target coverage. On the \textbf{middle} panel, we plot the fraction of biographies retained for each method against the target coverage level. Finally, on the \textbf{right} panel, we plot the number of facts per biography retained, again against the target coverage level.}
    \label{fig:conformal_fs_additional}
\end{figure}

\paragraph{Conformal Prediction.} For \textsc{Bio-FActScore}, we observe that multivalid conformal methods \textit{do not} improve performance across subgroups. In Figure \ref{fig:conformal_fs}, we observe very little difference in mean coverage error across groups. We hypothesize, however, that this negative result again is due to the smaller dataset size. In this case, our number of examples is the number of biographies in the dataset (683), giving us a calibration set size of 546 and test set size of 137. Further dividing the calibration and test sets into subgroups, it is possible there could simply not be enough examples per group for the distribution on the calibration set to generalize to the test set. Comparing the left panels of Figures \ref{fig:conformal_fs_additional} to \ref{fig:conformal_nq_additional}, we also find that even when looking at marginal coverage, all methods perform worse (the lines deviate more from $y=x$), likely due again to the small calibration and test size.


\end{document}